Heaven’s Light Is Our Guide

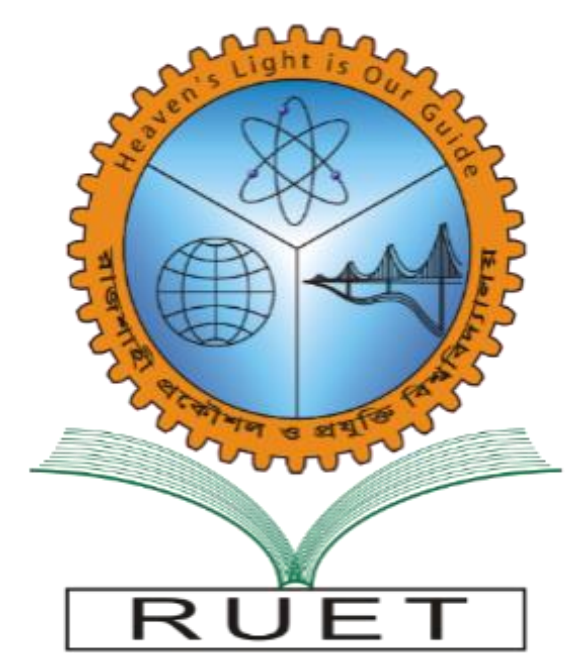

# Ensemble of Convolutional Neural Networks for Stroke Prediction: Towards Improved Diagnostic Accuracy

A Project/Thesis Submitted in Partial fulfillment for the requirement of the degree of

Bachelor of Science
in
Electrical & Computer Engineering

by

**MD. Shahriar Sajid**
**Roll No. 1810022**

Department of Electrical & Computer Engineering
Rajshahi University of Engineering & Technology
April, 2024

# Acknowledgement

This thesis has been submitted to the Department of Electrical and Computer Engineering of Rajshahi University of Engineering & Technology (RUET), Rajshahi-6204, Bangladesh, for the partial fulfilment of the requirements for the degree of B.Sc. in Electrical & Computer Engineering. Thesis title regards to "**Ensemble of Convolutional Neural Networks for Stroke Prediction: Towards Improved Diagnostic Accuracy**".

In the name of Allah, the most gracious and the most merciful. First and foremost, I am thankful to Almighty ALLAH for giving me the strength, knowledge, ability, and opportunity to undertake this study and complete it satisfactorily.

Secondly, I offer my sincere gratitude to my respected teacher and supervisor, Oishi Jyoti, Lecturer, Department of Electrical & Computer Engineering who has supported me throughout my industrial training with his patience and knowledge. I shall ever remain grateful to him for his valuable guidance, advice, and encouragement during my training.

I wish to thank once again Sagor Chandro Bakchy as the Head of the Department of Electrical & Computer Engineering for his support and encouragement and for providing all kind of laboratory facilities.

I would like to express my deep gratitude to all other respected faculty members of Department of Electrical & Computer Engineering, Rajshahi University of Engineering & Technology, for their inspiration, endless facilitation, and suggestions.

Finally, I wish to extend my sincere appreciation to my loving parents and my well-wishers, for their invaluable blessings and unwavering support. Additionally, I am grateful to my fellow classmates for their cooperation throughout this endeavor.

MD. Shahriar Sajid — April, 2024

Roll No. 1810022 — RUET, Rajshahi

# CERTIFICATE

This is to certify that the thesis entitled "**Ensemble of Convolutional Neural Networks for Stroke Prediction: Towards Improved Diagnostic Accuracy**" by MD. Shahriar Sajid (Roll No. 1810022) has been carried out under my direct supervision. To the best of my knowledge, this thesis is an original one and has not been submitted anywhere for any degree or diploma.

**Thesis Supervisor:**

.......................................

**Oishi Jyoti**

Lecturer

Department of Electrical & Computer Engineering

Rajshahi University of Engineering & Technology

# CERTIFICATE

This is to certify that the thesis entitled "**Ensemble of Convolutional Neural Networks for Stroke Prediction: Towards Improved Diagnostic Accuracy**" by MD. Shahriar Sajid (Roll No. 1810022) has been corrected according to my suggestion and guidance as an external. The quality of the thesis is satisfactory.

**External Member:**

.......................................

Department of Electrical & Computer Engineering

Rajshahi University of Engineering & Technology

# ABSTRACT

Brain stroke is characterized by high mortality and incidence rates, making it one of the most prevalent cases globally. It is well known for its serious damage to the human body particularly to the brain and is a rapidly growing disease for which the patient only gets a few hours to survive if they are affected by brain stroke. Early diagnosis and proper prevention measures can eradicate or minimize the percentage of life loss or permanent physical damages. Although brain stroke is lethal in most of the cases, proper assumption according to a patient's lifestyle and taking care of key medical terms can be life saving for a person who is about to get affected by brain stroke. In recent years, there has been a significant increase in novel computer-aided diagnostic techniques utilizing deep learning algorithms for the early detection of brain strokes. Predicting stroke effects from a set of predictive attributes may classify high-risk patients and guide cure approaches, leading to reduced relative incidence. In this work, an intelligent system was proposed that can make an effective prediction of possible brain strokes using only eleven (11) features. Seven (7) well known supervised machine learning algorithms were brought into action for the evaluation process. The overall process can be categorized into four phases. Phase 1: A comprehensive literature review of the various machine learning models, Phase 2: Brief discussion on the dataset, exploration and visualization, Phase 3: Preprocess the existing data and prepare for fitting in the neural network models. Phase 4: Finally feed the data to the appropriate models, obtain results and comment on them compared with each other. Notably, ensemble methods such as Random Forest, Stacking Classifier, and Bagging Classifier demonstrated remarkable performance, achieving accuracies of 99.52%, 99.52%, and 99.27% respectively. Additionally, Decision Tree exhibited strong classification accuracy at 98.24%. While KNN and TabNet Classifier achieved slightly lower accuracies at 96.73% and 96.49% respectively, they still showcased reliable performance. Our custom Feedforward Model achieved an accuracy of 94.91%, indicating its effectiveness in classification tasks. However, SVC and Logistic Regression exhibited comparatively lower accuracies at 88.06% and 77.03% respectively. Overall, our findings underscore the efficacy of ensemble methods and highlight their potential for accurate classification in similar datasets.

# Contents

# List of Figures

# List of Tables

# List of Abbreviations

| Abbreviation | Full Form |
|---|---|
| AI | Artificial Intelligence |
| CNN | Convolutional Neural Networks |
| ROC | Receiver Operating Characteristic |
| AUC | Area under the ROC Curve |
| RF | Random Forest |
| KNN | K Nearest Neighbours |
| LR | Logistic Regression |
| TabNet | Attentive Interpretable Tabular Learning |

# Chapter 1

# Introduction

## 1.1 Overview

This chapter briefly discusses about the whole process of the thesis which includes background and motivation, review of different authors' work, research gap on this topic, the main objective of the research, the thesis plan, estimation, and the thesis outline.

## 1.2 Background and Motivation

As to the latest data released by the World Health Organization (WHO), strokes rank as the second leading cause of mortality and the third leading cause of disability [1]. Stroke deaths accounted for 16.27% of all deaths in Bangladesh, placing the country at number 34 in the world. In Bangladesh, stroke ranks as the leading cause of death as well [2]. Therefore, early stroke prediction and identification, which will be useful for stroke detection and medication require the use of an expert system.

The rates of stroke-related mortality and disability are extremely high in underdeveloped countries. Roughly 75% of stroke-related deaths and 81% of stroke-related disability worldwide happen in developing countries [3]. Diseases both contagious and non-infectious pose a hazard to Bangladesh's sizable and expanding population. Stroke ranks third in Bangladesh in terms of causes of death, behind heart disease and infectious disorders like pneumonia and the flu. In Bangladesh, stroke-related deaths accounted for 134,166 fatalities in 2020, or 18.74% of all deaths, according to the latest WHO data. Bangladesh ranks 41st in the world with an age-adjusted mortality rate of 119.20 per 100,000 people [4].

To design and enhance an early prediction system that can assist physicians in the treatment process, machine learning techniques can be utilized to uncover hidden patterns in patient datasets. Machine learning algorithms in data science are rapidly becoming a vital tool for the diagnosis, management, and prognosis of various diseases.

## 1.3 Literature review

### 1.3.1 Analyzing Oznur Ozaltin et al. Work

For this purpose, the authors Oznur Ozaltin et al. [8] created the CNN architecture OzNet. For a hybrid method, they integrated OzNet with several machine learning algorithms (mRMR, Decision Tree, k-Nearest Neighbors, etc.). The model employs machine learning for categorization after extracting features from OzNet's fully connected layer. Actual CT scans of the brain were utilized to assess how well the suggested framework worked. When it came to stroke identification in brain CT scans, the OzNet-mRMR-NB (Naïve Bayes) combination had the best accuracy (98.42%) and Area Under the Curve (AUC) of 0.99. The paper doesn't explicitly mention how the study addressed potential biases in the training data.

### 1.3.2 Analyzing E. Dritsas et al. Work

Numerous machine learning algorithms were used by the work of E. Dritsas et al. [9], such as logistic regression, decision trees, random forests, K-Nearest Neighbors, support vector machines, and naive Bayes, were used by the researchers. They most likely made use of a dataset that included pertinent data about the patient's lifestyle, medical history, and demographics. Metrics like accuracy, sensitivity, specificity, and Area Under the Receiver Operating Characteristic Curve (AUC) were used to assess the model's performance. In comparison to conventional techniques, the study's findings indicate that machine learning algorithms, like Naive Bayes, which achieved 82% accuracy in this study, can be useful for predicting the risk of stroke. The stated accuracy of 82% might not be adequate for clinical decision-making in the actual world.

### 1.3.3 Analyzing Bhagyashree Rajendra Gaidhani et al. Work

The core approach applied by Bhagyashree Rajendra Gaidhani et al.’s work [10] revolves around CNNs and deep learning models for stroke detection. This suggests the paper might explore the design and training of these models for identifying strokes in brain scans. They utilized LeNet for classification of MRI images into normal and abnormal, and SegNet for semantic segmentation of abnormal regions and achieved high accuracy rates, with classification models reaching 96-97% and segmentation models achieving 85-87%.

### 1.3.4 Analyzing T. Tazin et al. Work

T. Tazin et al. [11] utilized the open-access Stroke Prediction dataset, applying Logistic Regression, Decision Tree Classification, Random Forest Classification, and Voting Classifier for model training. The Random Forest algorithm outperformed other models scoring 96% of accuracy, demonstrating robustness and reliability in stroke prediction. The study emphasizes the potential of ML in early stroke detection and intervention.

### 1.3.5 Analyzing G. Sailasya et al. Work

The study conducted by G. Sailasya et al. [12] aims to predict the occurrence of strokes using machine learning (ML) algorithms. The researchers employed various ML classification algorithms, including Naïve Bayes, Decision Tree, Random Forest, and Support Vector Machine (SVM). Naïve Bayes achieved the highest accuracy (around 82%) in stroke prediction. The paper suggests exploring neural networks and incorporating real-time brain images for improved stroke prediction.

### 1.3.6 Analyzing N. Biswas et al. Work

N. Biswas et al. [13] used in total eleven classifiers, including Support Vector Machine (SVM) and Random Forest, were analyzed. The classifiers showed over 90% accuracy before data balancing, and four displayed over 96% accuracy after using the oversampling method. The Random Over Sampling (ROS) technique was employed to balance the data. Performance metrics like Accuracy, F1-Measure, Precision, and Recall were used to evaluate the models. It was upon optimization that SVM scored 99.99% therefore recording the highest accuracy, as Random Forest came in second with an accuracy of 99.87%. Moreover, development of the most accurate model led to the creation of user-friendly mobile and web applications.

### 1.3.7 Analyzing B. Akter et al. Work

B. Akter et al. [14]'s model building process involved application of data collection, pre-processing, and transformation techniques on a 'brain stroke dataset' [15] using Random Forest (RF), Support Vector Machine (SVM), and Decision Tree (DT) classifiers, with data normalization through standardization during training and testing. The highest accuracy was recorded by the RF classifier at 95.30%. The paper further conducts a performance evaluation of the classifiers with performance measures such as accuracy, sensitivity, error rate among others.

### 1.3.8 Analyzing M. U. Emon et al. Work

Emon and his colleagues mainly conduct research on gauging different machine learning models which serve the purpose of forecasting strokes, one of the crucial aspects of medical diagnosis. In this work, the researchers have employed ten classifiers to predict strokes using Kaggle's public data set. Some of these classification algorithms are, Logistic Regression, Stochastic Gradient Descent (SGD), Decision Tree (DT), AdaBoost (AB), Gaussian Naive Bayes (NB), Quadratic Discriminant Analysis (QDA), Multi-layer Perceptron (MLP), KNeighbors (KNN), Gradient Boosting Machine (GBM), and Extreme Gradient Boosting (XGBoost). After that, the diverse original classifiers' outputs are combined by using weighted voting in order to enhance the performance according to accuracy rates. Furthermore, 97% accuracy concerning prediction rate has been reported for proposed study; outperforming all its rivals (the base classifiers) is weighted voting classifier.

### 1.3.9 Analyzing A. Srinivas et al. Work

A model was proposed by A. Srinivas et al. [17] which amalgamates a number of classifiers including Random Forest, Extremely Randomized Trees and Histogram-Based Gradient Boosting to create an ensemble algorithm. The ultimate prediction is a combination of predictions from all classifiers using soft voting; an average probability-weighted sum determines a decision point. This approach gave accuracies up to 96.88% on a Stroke prediction dataset at UCI machine learning repository [18]. A better type of classification could incorporate swarming optimization with intelligence.

### 1.3.10 Analyzing T. I. Shoily et al. Work

The possibility of machine learning for stroke detection is examined in T.I Shoily et al.'s [19] study. They use an unidentified patient data set to examine a number of methods, such as Random Forest, J48 decision trees, k-Nearest Neighbors, and Naive Bayes. The lack of information regarding the dataset and associated biases restricts the generalizability of the findings, even if the research probably offers the best accurate algorithm for this specific set of data. The classification procedure was carried out using a publicly available dataset from Kaggle. The classification procedure was carried out using four classifiers in all, of which Random Forest did the best, scoring an accuracy of 99.8%.

## 1.4 Research Gap

Although the stroke prediction dataset in Kaggle provides a valuable starting point for developing CNN models for stroke prediction, more research needs to explore the gap One major gap lies in the power limitations of the dataset itself. Long data sets can be a barrier to training exceptionally complex deep learning styles, which can hinder their ability to capture strong patterns in cases Besides, power biases will occur if estimates sequentially or linearly, especially those styles that perform well on accurate data sets but generalize poorly to real-world scenarios with many affected individual populations. In any case, the data set contains a higher proportion of healthy psychoanalysis compared to trauma-related data. This imbalance may limit the analytical power of the paradigm and further research is needed to investigate strategies to mitigate the limitations that should be created in stroke, without question that it is central to the neglected infection, and to ensure robust tumor prognoses.

## 1.5 Objectives of the Research

Some research and artificial intelligence that can prevent strokes already, these are survival situations in humans. Its framework is a heuristic neural network (CNN) that aims to use open data to predict shock events. “Early classification of injuries is important because they are sudden and the earlier treatment begins, the better the chances of full recovery” so the approach must be carefully planned because they are able to use for time-sensitive features such as classifying text or images.

The objectives are as follows:

1. Collect valid data from public resources.

2. Develop a sustainable algorithm that helps us to predict early occurrences of brain strokes.
3. Analyze the data using proposed methods and feed to the models.
4. Calculate the accuracy in each case and evaluate the prediction performance with other method's performances by using different evaluation matrices.

## 1.6 Vision

Convolutional Neural Networks (CNNs) have revolutionized industries because of their unique skills in visible records analysis. Their machine, driven through the human visible machine, permits them to extract significant features from photographs and motion pictures. This permits them to excel in tasks which include picture segmentation, object recognition, and facial popularity. As studies on CNNs maintain to conform, we will count on greater sophisticated packages to emerge, further transforming industries and shaping the destiny of computer vision.

## 1.7 Application Scenarios

Convolutional Neural Networks (CNNs) have revolutionized various fields by enabling advanced image and video analysis. Their ability to extract features and learn patterns from visual data has led to significant advancements in:

**Healthcare:** CNNs analyze medical scans (X-rays, MRIs) to assist in disease diagnosis and treatment planning. For example, they can detect anomalies in mammograms for early breast cancer detection [5].

**Mobile Devices:** CNNs enable features like facial recognition for secure unlocking and image segmentation for background removal in portrait photography on smartphones [6].

**Financial Sector:** CNNs analyze financial documents and transactions to detect fraudulent activities with greater accuracy, improving financial security [7].

These are just a few examples, and as research progresses, we can expect even broader applications of CNNs, further transforming various industries and shaping the future of computer vision.

## 1.8 Challenges

Convolutional Neural Networks (CNNs), a class of deep learning models, are instrumental in processing grid-like data such as images. However, achieving optimal performance with CNNs presents unique challenges.

1. **Requirement for Large Labeled Datasets:** CNNs typically require large amounts of labeled training data to learn effectively. The process of acquiring such vast datasets can be a daunting task. It often involves collecting numerous examples that accurately represent the problem space. Furthermore, each of these examples must be correctly labeled, which can be a labor-intensive and time-consuming process. The cost associated with data collection and labeling can also be significant, adding to the challenges.

2. **Susceptibility to Overfitting:** Overfitting is a common problem in machine learning, and CNNs are no exception. This issue becomes particularly pronounced when dealing with limited datasets. Overfitting occurs when a model learns the training data too well, to the point where it captures noise or details that are not representative of the underlying problem.

3. **Mitigation Techniques:** Several techniques are commonly employed to mitigate overfitting in CNNs. One such technique is data augmentation, which involves artificially expanding the training dataset by creating modified versions of the existing examples. This can include operations like rotation, scaling, or cropping in the context of image data. Another technique is regularization, which adds a penalty to the loss function to discourage complex models that are likely to overfit. Dropout layers, which randomly ignore a subset of features during training, can also be used to reduce overfitting by promoting robustness and preventing the model from relying too heavily on any single feature.

In conclusion, while CNNs are powerful tools in the field of deep learning, they come with their own set of challenges. Addressing these issues requires careful data collection and labeling, as well as the application of specific techniques to prevent overfitting.

## 1.9 Contributions

The fundamental contributions of our thesis work are as follows.

- The dataset was collected from a public repository named as "Stroke-prediction-dataset" which is a collection of information taken from the hospitals world-wide.
- We have provided a list of related works, where we summarized several works related to Machine Learning algorithms.
- We have reduced the feature by using feature importance score.
- We have proposed an algorithm and compared the performance of that algorithm with various Machine Learning algorithms.

## 1.10 Thesis Plan

### 1.10.1 Work plan

The process of early detection of brain stroke is a systematic and meticulous procedure that is divided into three integral parts: data collection, data processing, and modelling.

1. **Data Collection:** This is the initial phase where relevant data is gathered. The data, which forms the foundation of the entire process, is primarily acquired from public resources. These resources could include medical databases, health records, or research publications that are accessible to the public. The quality and accuracy of this data are crucial as they directly impact the subsequent stages of the process.
2. **Data Processing:** Once the data is collected, it undergoes a rigorous process of analysis. This stage involves scrutinizing the data to extract notable insights and identify patterns that could indicate a potential brain stroke. Advanced analytical techniques and algorithms are employed to process the data effectively. This stage is critical as the insights derived here form the basis for the modelling phase.
3. **Modelling:** The final stage involves creating predictive models based on the insights obtained from the data processing stage. These models are designed to predict the likelihood of a brain stroke. They are continually refined and tested for accuracy, ensuring they provide reliable predictions.

### 1.10.2 Timeline Diagram

Below section discusses the timeline of the thesis.

**June 2023:**

Started learning about dataset, python libraries and basic block of codes in Jupyter Notebook.

**August 2023:**

Gather knowledge about various types of brain stroke datasets and started literature review.

**September 2023:**

Select the stroke-prediction-dataset as the base dataset and started building a firm knowledge base on it.

**October 2023:**

Done with literature review. The review was done in two phases, one was the generic studies on brain stroke detection with both numerical and image dataset. The second one was strictly based on the selected stroke prediction dataset.

**November 2023:**

Form a detailed workplan which includes end to end processed of the research and start of building the code base.

**December 2023:**

Started cleaning the data and preprocess for feeding to the machine learning models.

**January 2024:**

Wrap up the code and submit it to supervisor for review.

**February 2024:**

Start writing the thesis book after getting the instructions from the supervisor.

**March 2024:**

Submit a draft to supervisor for correction and improvements.

**April 2024:**

Submit the finalized copy of the thesis book.

## 1.11 Thesis Estimation

### 1.11.1 Overall Estimation

This thesis research was conducted utilizing personal computing resources and the free tier offered by Kaggle. While this approach limited access to potentially more powerful computing resources, it demonstrates the feasibility of achieving valuable results using readily available tools. The free tier on Kaggle provided access to a vast amount of public data, which was crucial for training and evaluating the machine learning models explored in this thesis.

## 1.12 Thesis Outline

In the forthcoming chapters, we will delve into various pertinent investigations and insights related to our thesis. Our discussion will encompass:

**Chapter 2 - Methodology:** This chapter describes the case study of a base paper with its detailed methodology and literature review. Then justification to our study on the base paper.

**Chapter 3 – Design and Implementation**: This chapter employs the full overview of the design; details of the hard wares have been used and implementation of the proposed work. The implementation part covers dataset selection-description, data preprocessing, proper model selection with their description and lastly implementation of the models.

**Chapter 4 – Results and Overview:** This chapter encompasses the evaluation of the proposed methodology and a thorough comparison between the proposed methodologies.

**Chapter 5 - Social and Environmental Influence**: This chapter of the thesis discusses on the social and environmental influences of this study which includes societal impact, financial and health influences, safety, legal and cultural issues etc. along with impact of environment.

**Chapter 6 – Complex Engineering Problems and Activities:** This chapter represents a tabular form of complex engineering problems along with the activities discussed in brief.

**Chapter 7 - Social and Environmental Influence:** This chapter of the thesis discusses on the social and environmental influences of this study which includes societal impact, financial and health influences, safety, legal and cultural issues etc. along with impact of environment.

**Chapter 8 – Conclusion and Future work:** This chapter concludes the research work with a brief discussion on what has been achieved throughout the study and what are the possible future endeavors in this field.

# Chapter 2

# Methodology

## 2.1 Overview

This chapter will cover selection of base research among the reviewed papers, explanation of their methodology, case study of that research and finally justification will be given of our study.

## 2.2 Case study

For case study we have selected T. Tazin et al. [11] research among the other works because T. Tazin et al. employed the same dataset and similar machine learning models which we used in our research. Using the publicly available Stroke Prediction dataset, they trained their model using voting classifiers, decision trees, random forests, and logistic regression. With a 96% accuracy rate, the Random Forest algorithm outperformed other models, proving its robustness and dependability in stroke prediction. The study highlights machine learning's potential for early stroke detection and treatment.

## 2.3 Detailed Methodology

The employed methodology by T. Tazin et al. [11] is listed below.

**Dataset Selection:** They selected the stroke-detection-dataset from Kaggle, consisting of 5110 instances with 12 features.

**Preprocessing:** For preprocessing the employed dataset three step approach has been followed which are:

1. Missing data handling
2. Label encoding
3. Handling imbalanced data

Handling missing values, label encoding, and balancing the dataset SMOTE technique has been employed.

**Model Construction:** The authors here used machine learning algorithms like Logistic Regression, Decision Tree, Random Forest, and Voting Classifier for building predictive models on the employed dataset.

**Result Analysis:**

Among the four employed models, Random Forest outperformed the other machine learning models. The below table shows a basic overview of the results of the study.

| S/L No. | Model Name | Precision | Recall | F1 Score | Accuracy (%) |
|---|---|---|---|---|---|
| 1 | Random Forest | 0.97 | 0.97 | 0.96 | 96 |
| 2 | Voting Classifier | 0.91 | 0.91 | 0.91 | 91 |
| 3 | Logistic Regression | - | - | - | 79 |
| 4 | Decision Tree | 0.94 | 0.95 | 0.95 | 94 |

## 2.4 Justification of the study

For our study we have employed basic machine learning models along with some ensemble learning techniques. The ensemble techniques opened doors towards combining different types of models and obtaining a sustainable accuracy with 5-folding techniques. Here the reference paper used only four machine learning models and obtained the highest accuracy of 96%. We aim to preprocess the data in a bit more efficient way. Then employ the basic machine learning models followed by ensemble learning techniques like bagging and stacking classifier where we choose base model by voting among the employed models so that we can obtain a sustainable accuracy which should outperform the existing studies.

# Chapter 3

# Design and Implementation

## 3.1 Design overview

This chapter describes the design and implementation process of the thesis which includes the hardware details, dataset selection, dataset description, dataset preprocessing, model selection and description and finally model implementation on the processed dataset.

### 3.1.1 Hardware design

TABLE 1: DESCRIPTION OF SYSTEM CONFIGURATION (REMOTE)

| **System Configuration (Kaggle Notebook)** | |
|---|---|
| Processor | Intel Xeon CPU |
| CPU | ~2.20 GHz |
| RAM | 32 GB |
| GPU | NVIDIA A100 |
| GPU RAM | 40 GB |
| Hard Disk | 80 GB |

TABLE 2: DESCRIPTION OF SYSTEM CONFIGURATION (LOCAL)

| **System Configuration (Local Machine)** | |
|---|---|
| Processor | AMD Ryzen 5 5500 |
| CPU | 3.6 ~ 4.2 GHz |

| RAM | 16 GB |
|---|---|
| GPU | NVIDIA RTX 3060 ti |
| GPU RAM | 8 GB |
| Hard Disk | 1 TB |

## 3.2 Implementation

This section will offer a concise overview detailing the dataset's characteristics, demonstrating data visualization and analysis techniques and preprocessing techniques performed on the dataset before training it to the model.

### 3.2.1 Dataset selection

For the research a public dataset name stroke-prediction-dataset [15] from Kaggle has been used which is a numerical dataset having 5110 rows and 12 columns. Dataset detailed description will be discussed in the following chapters.

### 3.2.2 Description of the Dataset

The process of creating a predictive model begins with selecting a dataset. In this scenario, a Kaggle dataset is used, including various physiological features as attributes. These characteristics serve as the foundation for later analysis and final projection. The dataset has 5110 rows and 12 columns, providing a large amount of data for the model to learn from. The output column in this dataset is 'stroke', which has two possible values: 1 or 0. Several 0 indicates that no stroke risk was recognized, but a value of 1 indicates that a stroke risk was discovered. It's important to note that in this dataset, the probability of 0 in the output column (stroke) exceeds the possibility of 1. Specifically, 249 rows in the stroke column have the value 1, indicating stroke risk, whereas a significantly larger number, 4861 rows, have the value 0, indicating no identified stroke risk.

Before the dataset can be used for machine learning, it must first undergo a process known as Data Preprocessing. This involves cleaning the dataset and preparing it in a way that the machine learning model can understand. The first step in this process is to check for null values in the dataset and fill them in appropriately. Following this, Label Encoding is performed to convert string values into integers, making them more digestible for the model. If necessary, One Hot Encoding is also performed to further transform categorical variables.

It's worth noting that the dataset in question is unbalanced, meaning that the classes in the output column are not represented equally. This can pose a challenge as it may lead the model to be biased towards the majority class, thereby affecting its accuracy and precision. To overcome this, data balancing techniques must be employed to ensure that all classes are adequately represented.

Once the Data Preprocessing is complete, the dataset is split into two subsets: training data and testing data. The training data is used to build the model, while the testing data is used to evaluate its performance. A model is then built using this preprocessed data. Various Classification Algorithms are employed in this step, each offering a different approach to the task of prediction. The performance of each algorithm is evaluated by calculating its accuracy, which is the proportion of true results (both true positives and true negatives) among the total number of cases examined.

Finally, the accuracies of all the algorithms are compared to identify the best-trained model for prediction. This model, having demonstrated the highest accuracy, is then selected for making the final predictions.

### 3.2.3 Description of the attributes

There are 12 attributes in total in the dataset. The below table discusses about the dataset briefly for each attribute with their values and description.

TABLE 3: ATTRIBUTES OF THE STROKE PREDICTION DATASET

| **Attribute Name** | **Type** ***(Values)*** | **Description** |
|---|---|---|
| 1. id | Integer | A unique integer value for patients |
| 2. gender | String literal | Tells the gender of the patient |
|  | *(Male, Female, Other)* |  |
| 3. age | Integer | Age of the Patient |
| 4. hypertension | Integer *(1, 0)* | Tells whether the patient has hypertension or not |
| 5. heart_disease | Integer *(1, 0)* | Tells whether the patient has heart disease or not |
| 6. ever_married | String literal *(Yes, No)* | It tells whether the patient is married or not |
| 7. work_type | String literal *(children, Govt_job, Never_worked, Private, Selfemployed)* | It gives different categories for work |
| 8. Residence_type | String literal *(Urban, Rural)* | The patient's residence type is stored |
| 9. avg_glucose_level | Floating point number | Gives the value of average glucose level in blood |
| 10. bmi | Floating point number | Gives the value of the patient's Body Mass Index |
| 11. smoking_status | String literal *(formerly smoked, never smoked, smokes, unknown)* | It gives the smoking status of the patient |
| 12. stroke | Integer *(1, 0)* | Output column that gives the stroke status |

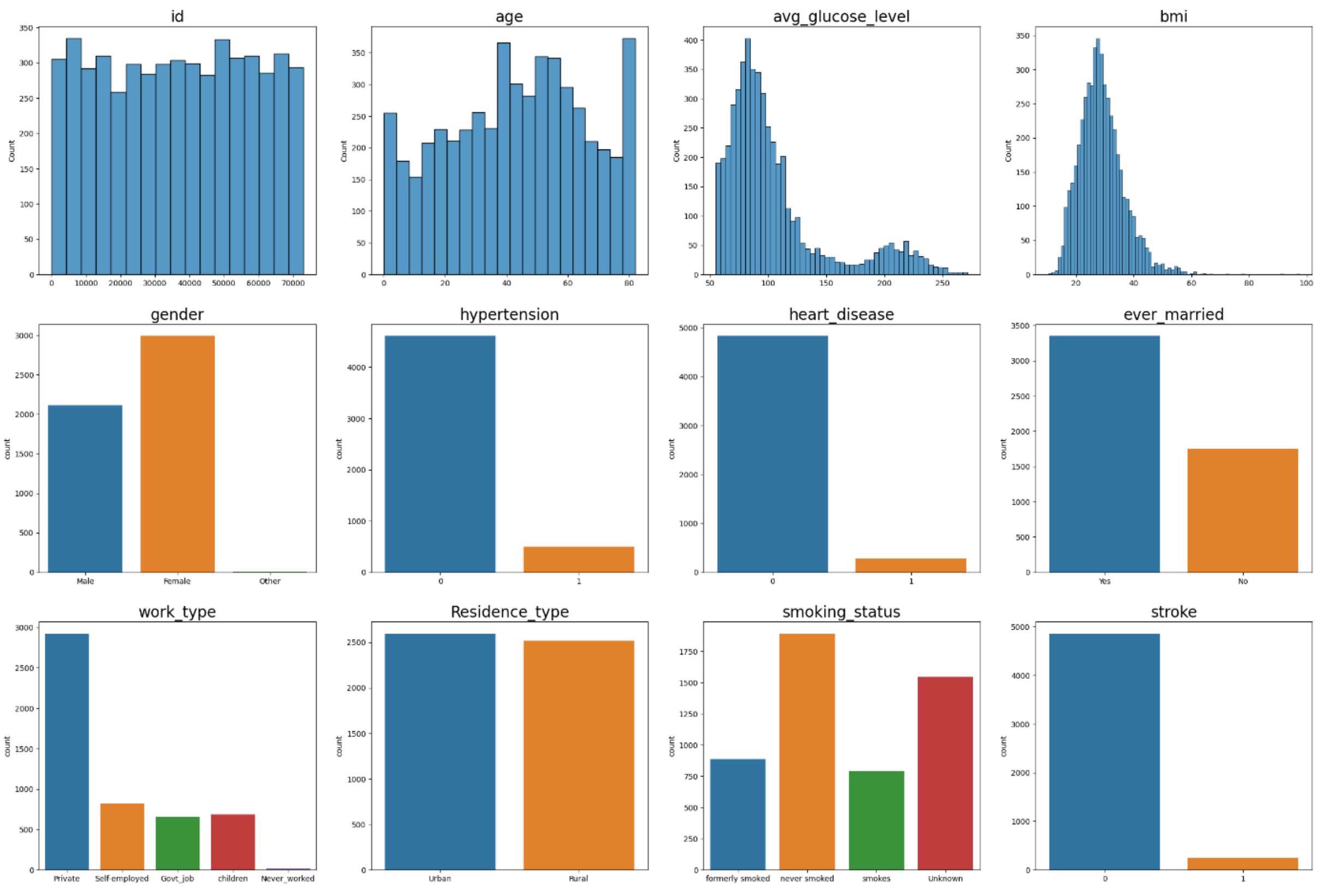

Figure 1: Visualizing the features of the stroke prediction dataset

### 3.2.4 Dataset Splitting

Splitting the dataset into a training set and a testing set is a crucial step in evaluating the performance of the model. The dataset is divided in an 80:20 ratio, where 80% of the data is used for training the model, allowing it to learn and identify patterns, and the remaining 20% is reserved for testing.

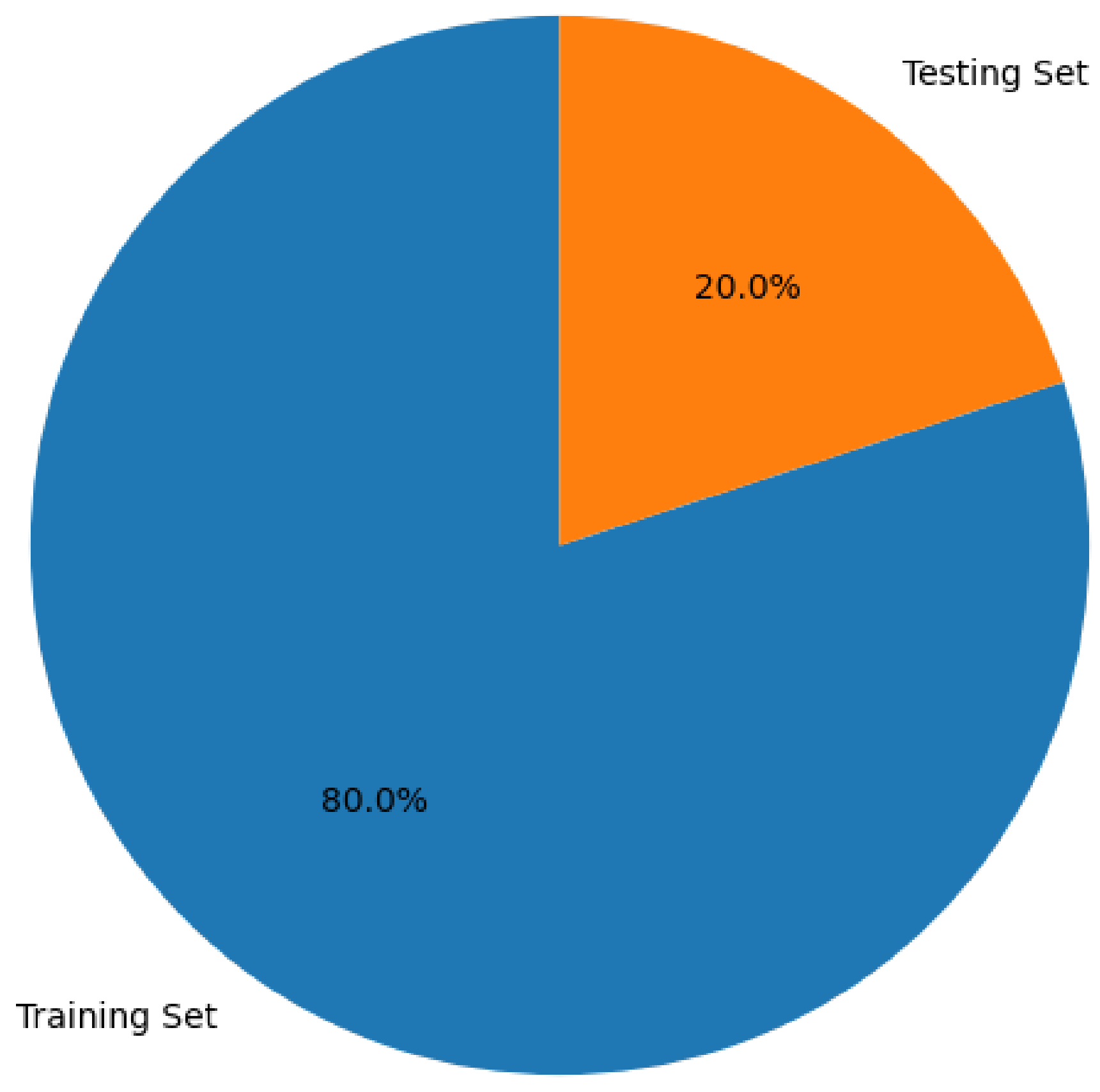

Figure 2: Test-train split of the employed dataset

This testing portion is crucial as it provides an unbiased evaluation of a model's performance, ensuring that the model has not just memorized the data, but has learned from it in a way that is generalizable to new, unseen data.

### 3.2.5 Dataset Preprocessing

Data Preprocessing is a crucial step in the machine learning pipeline. It is required before model building to ensure that the data is in the best possible form for the algorithms to learn from. This step involves cleaning the data by removing unwanted noise and outliers, which could otherwise lead to deviations in the model's training process and ultimately affect its performance. Anything that could potentially hinder the model's efficiency is addressed during this stage.

After the appropriate dataset has been collected, the next step is to prepare it for model building. This involves a series of tasks aimed at refining the data and making it suitable for the machine learning model. The stroke prediction dataset comprises 12 attributes, each representing a different aspect of the data. The first step in the preprocessing stage is to drop irrelevant columns from the dataset. In this case, the 'id' column is dropped because it does not contribute significantly to the model building process. The 'id' values are unique to each entry and do not carry any meaningful information that could help in predicting the outcome.

Following this, the dataset is checked for null or missing values. Null values in a dataset can be problematic as they can lead to errors during the model training process and can also skew the results. Therefore, it is essential to handle these appropriately. In this dataset, the 'bmi' column was found to have null values. These were filled with the meaning of the column data. Using the mean is a common strategy for handling null values as it does not drastically affect the overall distribution of values in the column.

In conclusion, Data Preprocessing is a vital step in the machine learning process. It involves cleaning the data and transforming it into a form that the machine learning model can easily understand. By doing so, it ensures that the model is trained in high-quality data, thereby improving its performance and accuracy.

### 3.2.6 Handling Imbalanced Data

The dataset used for stroke prediction is severely unbalanced. The dataset contains 5110 rows, with 249 rows indicating the presence of a stroke and 4861 rows indicating the absence of a stroke. Figure 2 shows a graphical illustration of the imbalance. Training a machine-level model on such data may result in accuracy, but other accuracy metrics such as precision and recall are limited. If such unbalanced data is not handled properly, the findings will be inaccurate and the forecast inefficient. As a result, to obtain an efficient model, the imbalanced data must be addressed first. For this reason, the under-sampling method is applied. Under sampling [21] balances the data wherein the majority class is under sampled to match the minority class. In this case, the class with a value as '0' is under sampled for the class with the value' 1'. So, after under sampling the resulting dataset will have 249 rows with value ‘0’ and 249 rows with value ‘1’. The graphical representation of the dataset before and after handling imbalanced data is shown in the figure below.

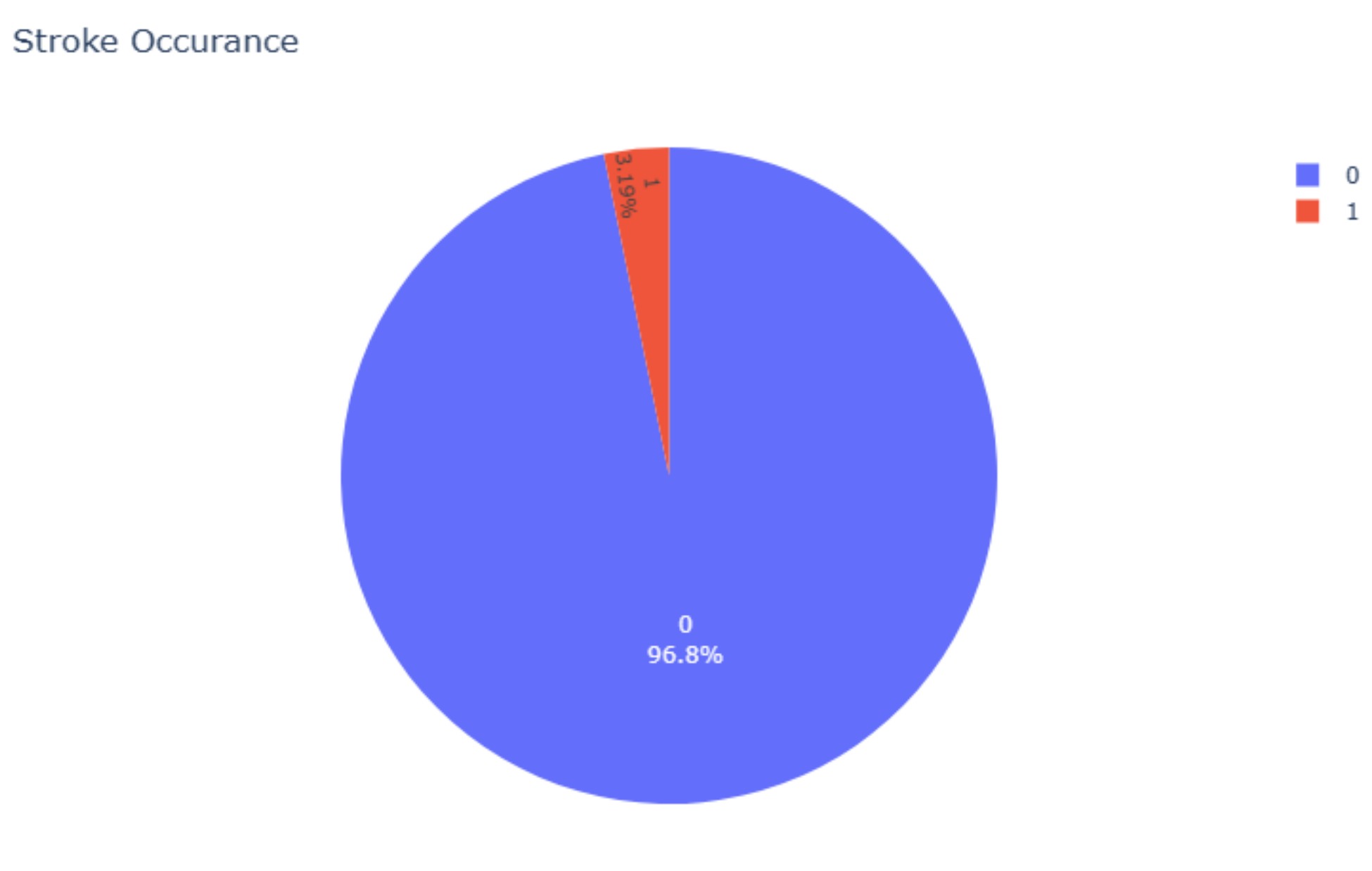

Figure 3: Before Sampling the data

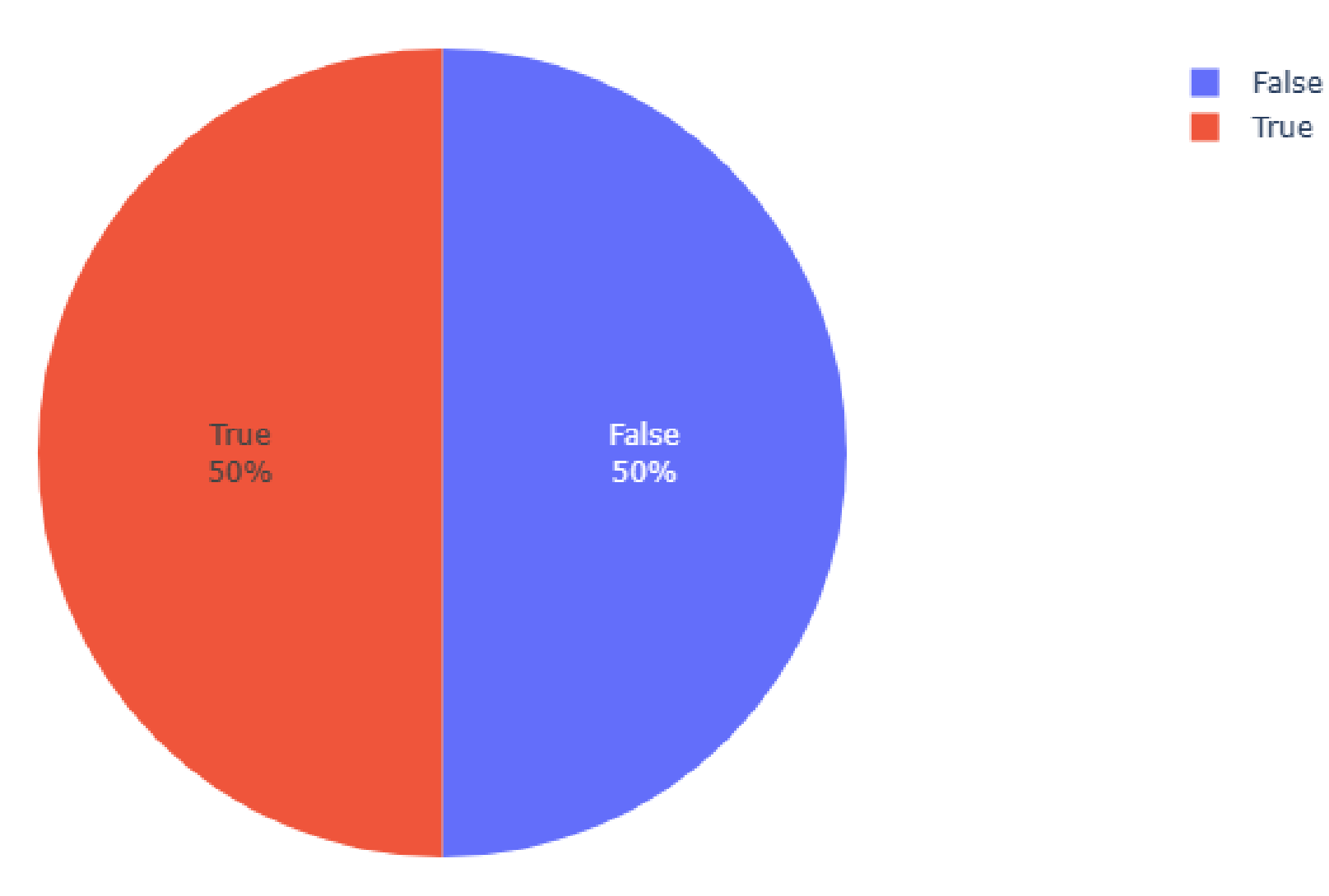

Figure 4: After Sampling the data

### 3.2.7 Machine learning architecture: CNN

Convolutional neural networks (CNNs) [22], or ConvNets for short, are a subset of deep neural networks that have been specially designed to perform exceptionally well in tasks involving grid-like data, like images and videos. In a variety of tasks, including image classification, object detection, facial recognition, and other related applications, they have proven to be highly effective. Convolutional Neural Nets (CNNs) process input data by applying a series of operations in a methodical manner. CNNs can learn hierarchical patterns and features on a range of scales on their own through this process.

The basic function at the center of a convolutional neural network (CNN) is convolution. Filters or kernels with learning capabilities make up convolutional layers. These filters, which are akin to tiny sliding windows, move across the input data, like an image. Features are extracted by performing element-wise multiplication and then adding the results in the receptive field for every filter. The previously mentioned process oversees producing feature maps that successfully highlight different patterns in the input.

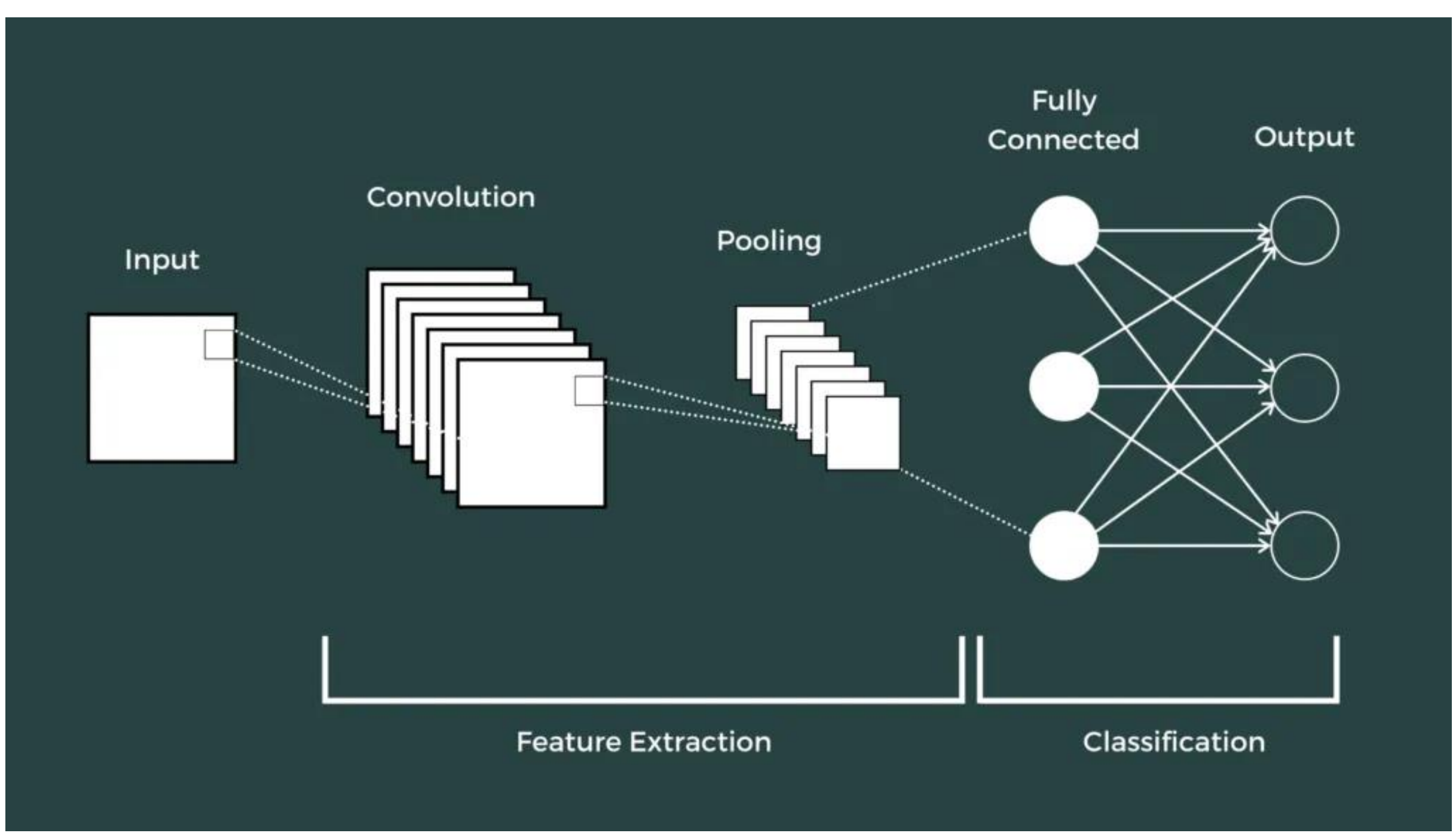

Figure 5: Simple CNN Architecture [23]

After the convolution process, a non-linear activation function known as ReLU (Rectified Linear Unit) is applied element-by-element to the feature maps. The model can better grasp intricate relationships and efficiently capture higher-level features thanks to the addition of non-linearity. One kind of layer that successfully reduces the spatial dimensions of feature maps without sacrificing important information is a pooling layer. To download sample feature maps, pooling operations are frequently used in a variety of applications, including computer vision. The pooling operations max-pooling and average-pooling are two that are frequently used. While average-pooling calculates the average value within these regions, max-pooling chooses the maximum value within small regions. Both processes successfully lower the feature maps' dimensionality.

The output obtained from the preceding pooling layer is converted into a vector with dimensions of one. The preparation of the data for fully connected layers is the responsibility of this specific step. These are similar layers to those in traditional feedforward neural networks. They perform high-level feature extraction and classification. Every neuron in a fully connected layer is connected to every other neuron in the layer above it, allowing the neural network to learn complex, broad patterns and produce predictions for various classes. When dealing with multi-class classification, the last fully connected layer typically has the same number of neurons as the total number of output classes.

The network's prediction for the input data is produced by the fully connected layer. The neural network uses a SoftMax function to produce a probability distribution over different classes when performing classification tasks. Finding the class in this distribution with the highest probability yields the predicted class.

### 3.2.8 Model Selection

The preprocessed data is fed into various machine learning models. The models listed in the flowchart are:

- Logistic Regression
- Support Vector Machine
- TabNet Classifier
- Custom Feedforward Model
- Decision Tree
- Bagging Classifier
- Random Forest
- Stacking Classifier

### 3.2.9 Model Description

### 3.2.10 Logistic Regression

A statistical method called logistic regression is used to examine a dataset that has one or more independent variables that determine the outcome. Two possible outcomes are found in the variable used to calculate the result. Finding the most appropriate model to explain the relationship between the variables is the aim of logistic regression. The coefficients of a formula that can forecast a logarithmic transformation of the likelihood of an interest characteristic occurring are produced by logistic regression.

$$\text{logint}(p) = a_0 + a_{1X_1} + a_{2X_2} + \cdots + a_{iX_i}$$

where p is the probability of presence of the characteristic of interest.

The logistic function, also known as the sigmoid function [25], is an S-shaped curve that maps any real-valued number into a value between 0 and 1. This value can be interpreted as the probability of the positive class. In other words, logistic regression models the relationship between predictor variables and a categorical dependent variable by estimating probabilities using this logistic function.

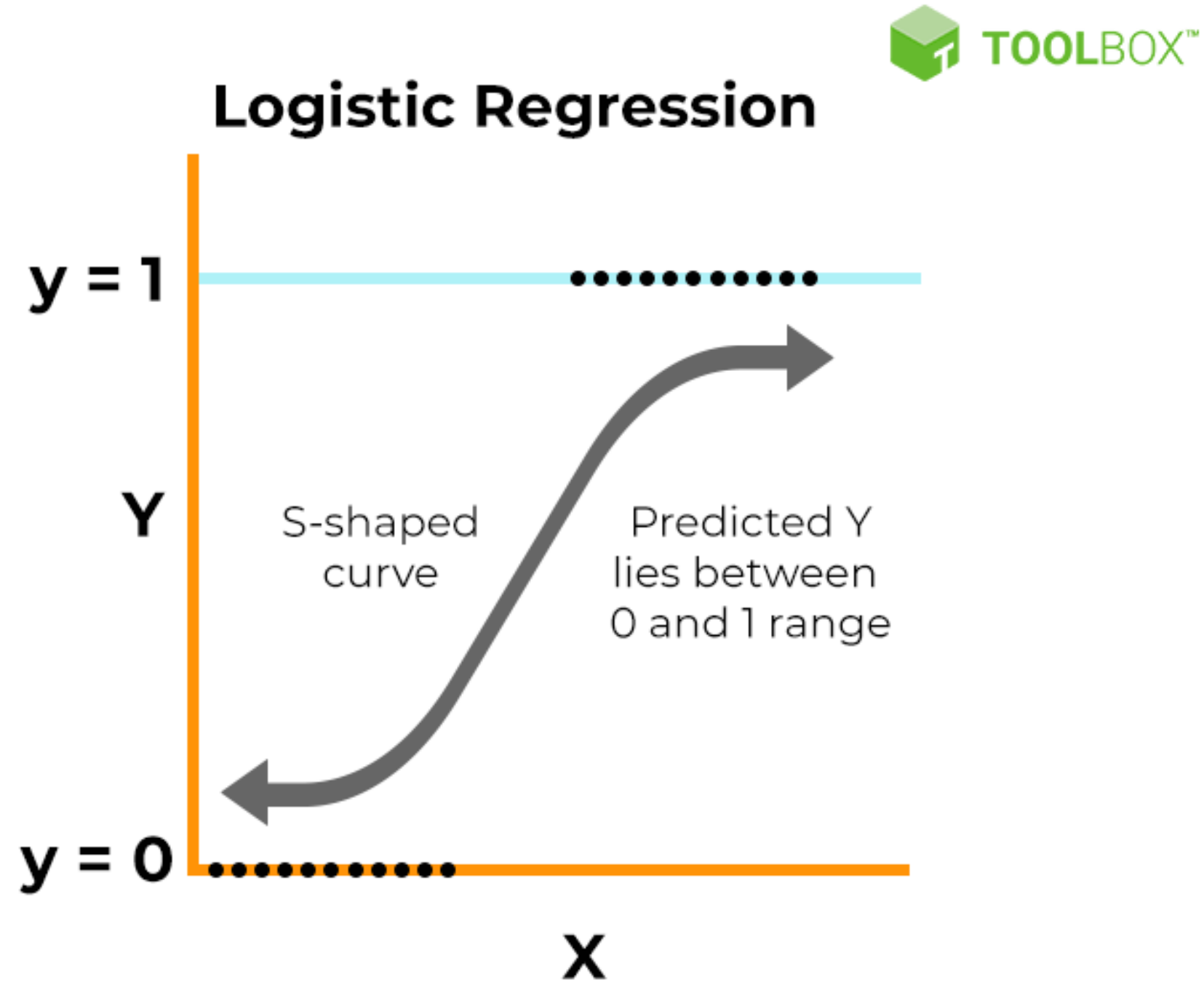

Figure 6: Simple logistic regression S-shaped Curve [24]

Despite its simplicity and interpretability, logistic regression can be extended to handle more complex scenarios. For example, it can be used for multiclass classification problems, where the outcome variable has more than two categories. In this case, a separate logistic regression model is fit for each category of the outcome variable against all others. This approach is known as one-vs-all (OVA) or one-vs-rest (OVR) logistic regression. Furthermore, logistic regression forms the basis for more advanced methods like neural networks. A single neuron in a neural network is essentially performing logistic regression.

### 3.2.11 Support Vector Machine

A Support Vector Machine (SVM) [26] is a supervised machine learning algorithm or discriminative classifier based on a splitting hyperplane. This hyper-plane is a line that divides a plane into two sections in two-dimensional spaces, with each class remaining on either side and separating the various classes of data.

At its core, SVM seeks to discover an ideal hyperplane in an N-dimensional space that successfully separates data points into different classes. The primary aim is to maximize the margin between these classes nearest points. This margin indicates the distance between the hyperplane and the nearest data points. By doing so, SVM determines the optimal separating hyperplane, often known as the maximum-margin hyperplane or hard margin.

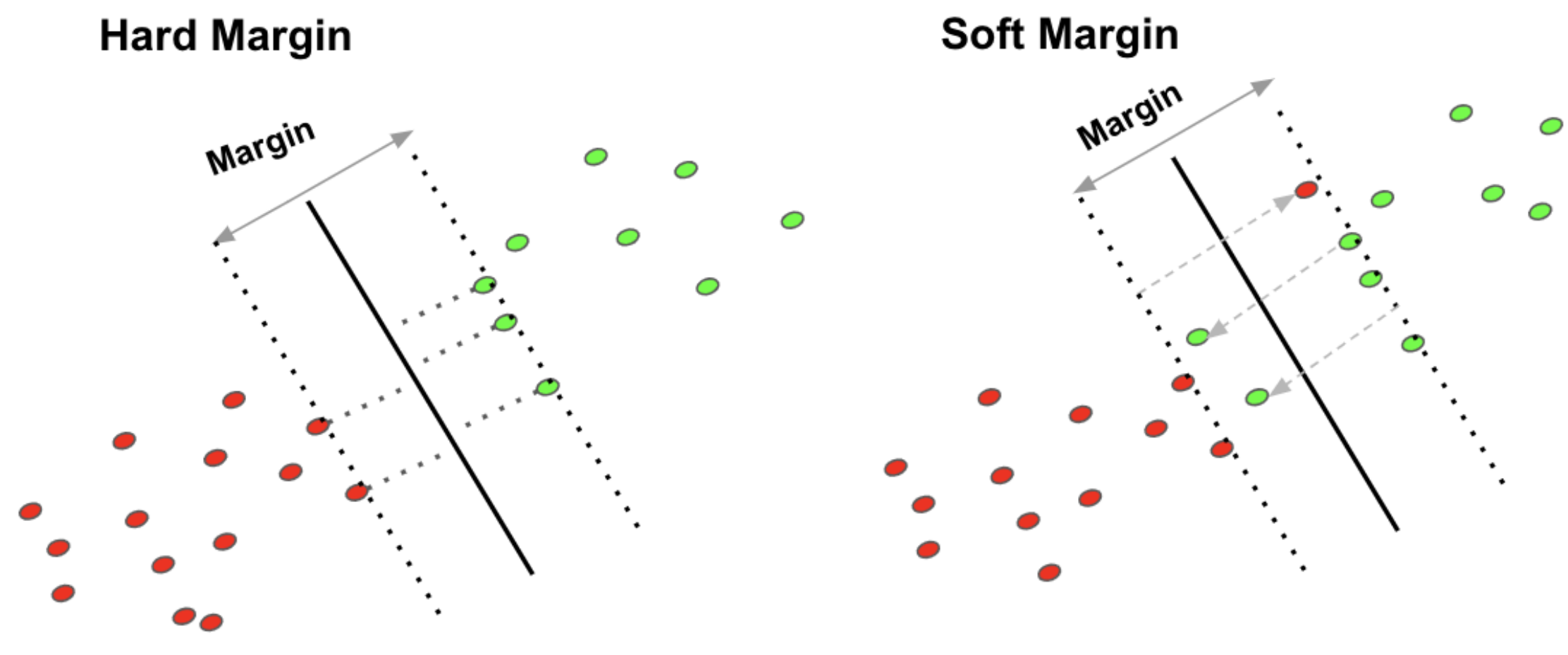

Figure 7: Hard margin and soft margin SVM visualization [27]

When data points are linearly separable, SVM aims to find a hyperplane that best separates the classes. The hyperplane is defined by a linear equation:

$$w^T x + b = 0$$

where (w) represents the weight vector (normal to the hyperplane), (x) is the input feature vector, and (b) is the bias term. The data points closest to the hyperplane are called support vectors. These support vectors influence the position and orientation of the hyperplane.

**Kernel Trick and Nonlinear Separation:**

In real-world scenarios, data is often not perfectly linearly separable.

SVM tackles this challenge by employing the kernel trick. Instead of directly mapping data to a higher-dimensional space, it uses a kernel function to implicitly transform the input features.

The kernel function computes the dot product between the transformed feature vectors in the higher-dimensional space. Commonly used kernels include:

**Polynomial Kernel:** It captures polynomial relationships between features. The equation becomes:

$$K\ (x, x') = (x^T x' + c)d$$

where (c) and (d) are hyperparameters.

**Radial Basis Function (RBF) Kernel:**

**Optimal Separation and Margin:** SVM aims to maximize the margin—the distance between the hyperplane and the nearest support vectors.

The optimal hyperplane is the one that maximizes this margin while ensuring correct classification.

SVM solves a convex optimization problem to find the optimal weights (w) and bias (b).

### 3.2.12 Decision Tree Classifier

Decision Trees (DTs) are a fundamental non-parametric supervised learning method utilized for both classification and regression tasks [28]. Renowned for their simplicity yet effectiveness, decision trees construct a hierarchical structure resembling a tree. This structure begins with a single node, often referred to as the root, which then branches out into multiple nodes, akin to the branches of a tree. Each internal node and the root represent a feature within the dataset, while each leaf node represents a distinct class label. The paths traversed from the root to the leaf nodes encapsulate the decision-making rules employed by the model for classification.

In the realm of machine learning, a decision tree embodies its name by adopting a tree-like structure to categorize data or instances. Commencing at the root node, which possesses no incoming edges, the decision-making process unfolds by traversing through the tree until reaching a leaf node, which typically has exactly one incoming edge. Integral components of a decision tree include decision nodes, leaf nodes, edges, and paths. Leveraging these elements, a decision tree effectively discerns patterns within input data or attributes to make informed decisions or classifications.

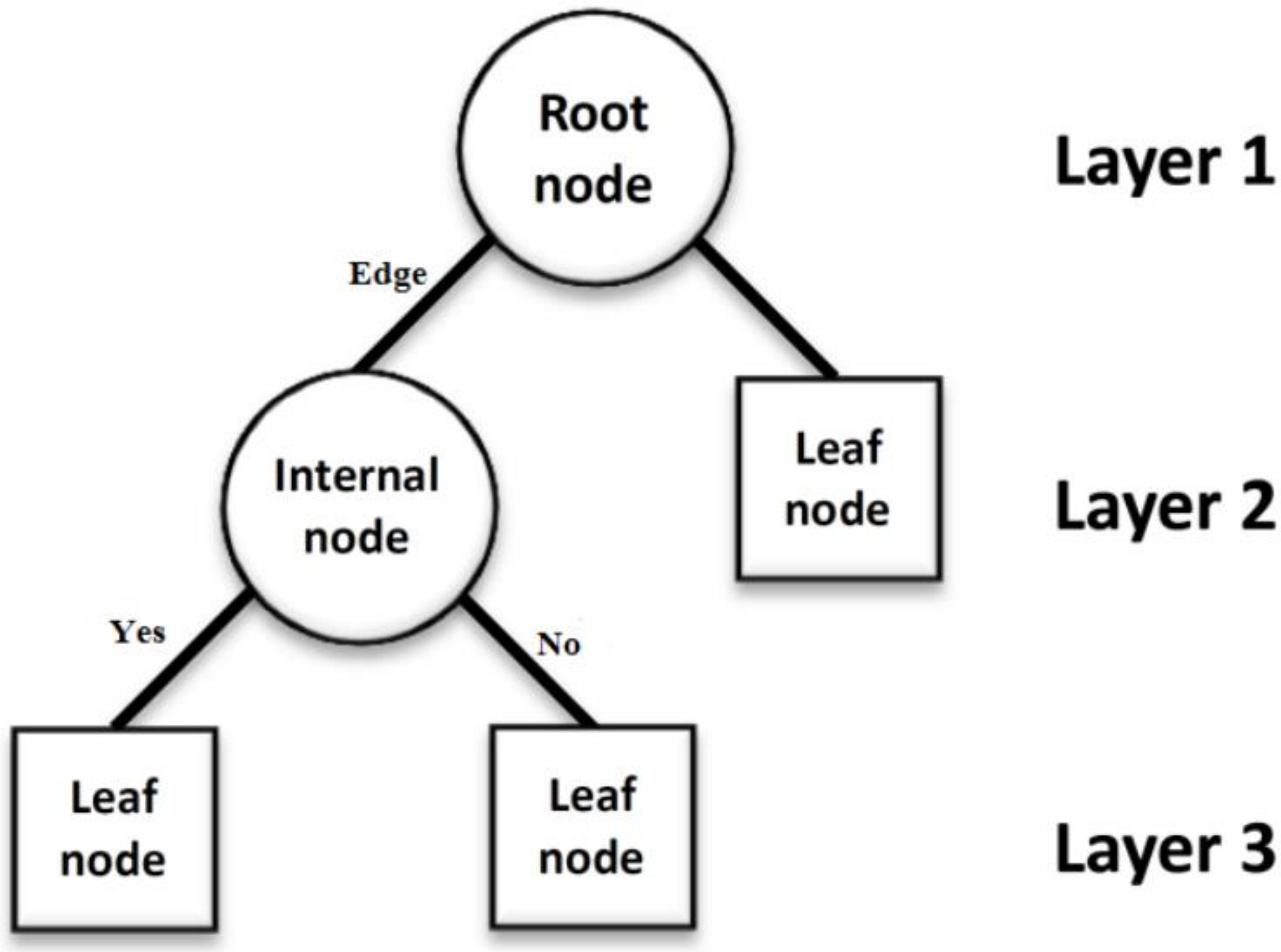

Figure 8: A simple construction of decision tree classifier

Decision trees serve as versatile and interpretable models in machine learning, offering insights into the decision-making process through their transparent structure. By delineating classification rules in a hierarchical manner, decision trees facilitate comprehensible interpretation and visualization of complex decision boundaries, making them invaluable tools in various domains, ranging from finance to healthcare and beyond.

### 3.2.13 Random Forest

In machine learning, random forests are a popular cluster learning technique. Decision forests, known for their robustness and flexibility, are decision trees that are voted or tossed to contribute to the final classification or regression results in an unordered forest system. As the day progresses, more decision trees are collected, lowering the risk of overfitting, and improving generalization performance.

To employ a random forest, several decision trees are built, each of which is trained with a unique set of characteristics. By randomly infusing diversity into the individual trees in the training data, this technique promotes collective decision making using the wisdom of the crowd.

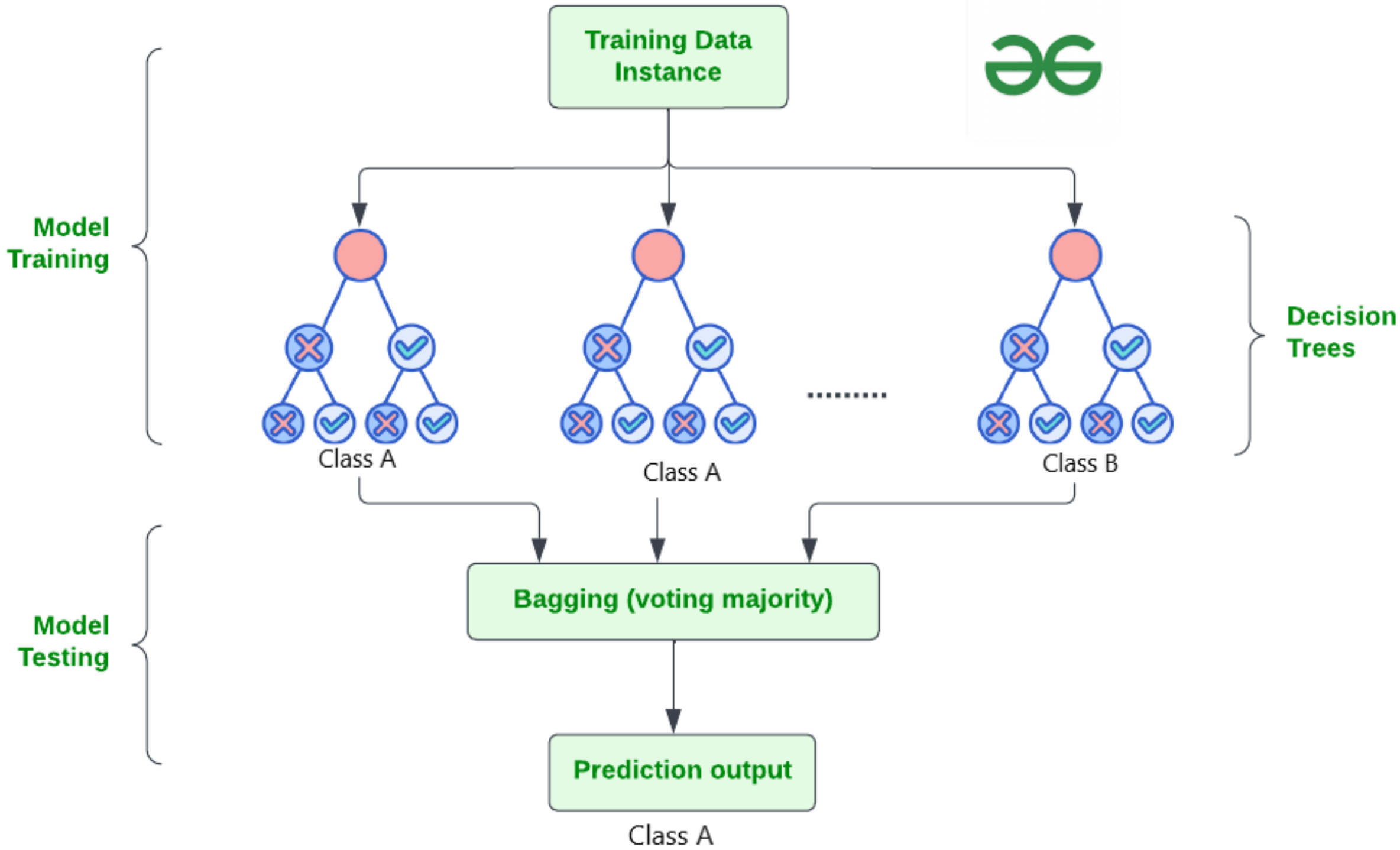

Figure 9: Visual representation of training-testing phase of random forest classifier [29]

The widespread adoption of open forests in various areas can be attributed to its many advantages. First, due to the composition of decision trees, randomness injected during tree construction and feature selection, random forest is robust to overfitting This robustness enables random forest to generalize well to unseen data, making it particularly suitable for datasets that are dimensional or superior noise characteristics and do not require extensive tuning, making it an attractive choice for practitioners looking for low-maintenance but effective machine learning models Furthermore, random forests do not exhibit natural interpretation, because they do not difficult to provide insights into feature importance based on the frequency of feature use in a tree And in addition to enable classification or identify key drivers of regression performance, random forest is highly scalable and capable handle large data sets efficiently, due to its parallel nature and ability to independently train each decision tree

In practical applications, random forests find extensive applications in a wide range of industries including healthcare, finance, commerce, bioinformatics, and it's feasible and robust capabilities make it a valuable diagnostic tool, credit risk analysis, customer segmentation, genetic analysis etc. In addition, random forests seamlessly integrate with other machine learning techniques, e.g., Ring, Dimensionality reduction and model stacking, further enhance its usefulness and efficiency.

### 3.2.14 Bagging Classifier

Bagging Classifier, short for Bootstrap Aggregating, is designed as an ensemble learning method to improve the stability and accuracy of machine learning models. This method is implemented by generating more subsets of the original dataset by replacing them with random sampling known as bootstrapping. Each subset is then used to train the base classifier independently. [30] Predictions from these base classifiers are collected using techniques such as averaging (for regression) or polling (for classification) to make a final prediction.

The main advantage of Bagging Classifier is its ability to reduce differences and reduce war Eligibility Overqualification. By performing multi-base classification training on different small data sets, bagging classifier introduces changes to the cluster, thereby reducing noise or Disadvantages include improved generalization performance. It can effectively handle datasets with high features and nonlinear decision boundaries and makes it so suitable for a wide range of machine learning projects.

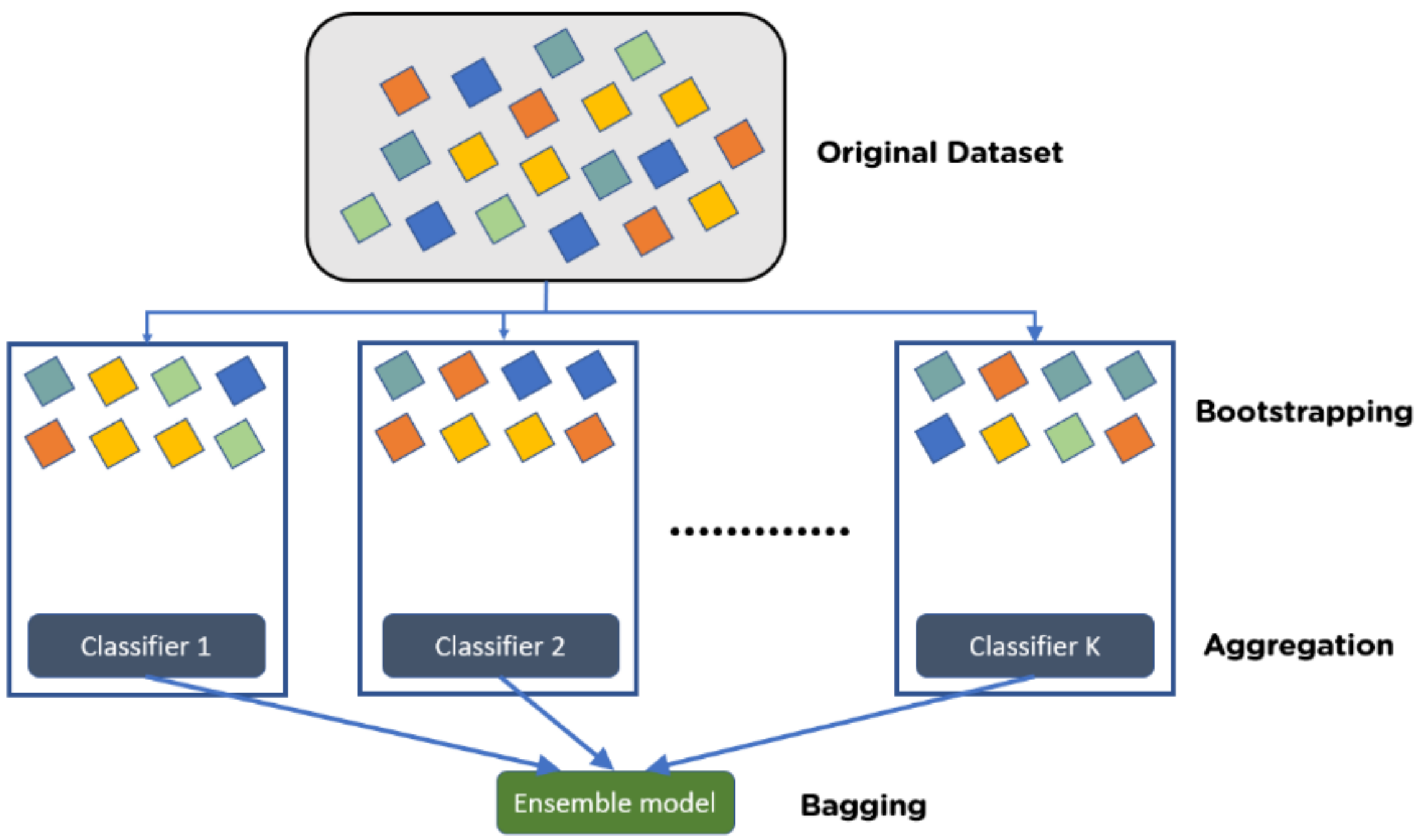

Figure 10: Simplified diagram of bagging ensemble technique [31]

Given, training dataset $Q = \{(x_1, y_1), (x_2, y_2) \ldots (x_M, y_M)\}$, where $y_M = \{0, 1\}$ corresponds to the class types (Non-stroke and Stroke). $B_S$ is the number of base models or subsets and $B_C$ is the base classifier.

1. Generate N number of bootstrap datasets as $Qn$ = Bootstrap($Q$), where

$n = 1, 2, \ldots, B_S$

2. Initialize the base model set $B_S = 0$

3. Create base model $b_{mn} = (Qn)$, where $n = 1, 2, \ldots, B_S$ by applying the base classifier $B_C$ on each bootstrap

4. For $n = 1, 2, \ldots, B_S$ do: $S = B_S \cup \{b_{mn}\}$

5. Result is given by bagging method is: $FBagging(x)$ = mostly voted class in $V_m(x)$, where $V_m \in B_S$ and $x$ is the predicted test pattern. [32]

In practice, Bagging Classifier has found widespread application across various domains, including but not limited to classification, regression, and anomaly detection. Its versatility, robustness, and ability to improve model performance make it a popular choice for tackling complex machine learning problems. Moreover, Bagging Classifier can be combined with other ensemble methods, such as Random Forest and Gradient Boosting, to further enhance predictive accuracy and stability. Overall, Bagging Classifier stands as a valuable tool in the machine learning toolkit, offering a reliable and scalable approach to ensemble learning.

### 3.2.15 Stacking Classifier

Stacking Classifier is a powerful ensemble getting to know technique that combines predictions of a couple of base classifiers using a meta-classifier to make a very last prediction Unlike traditional ensemble techniques which include bagging and boosting, stacking calls for base training exceptional styles of classifiers, every of which can use specific algorithms or parameter settings [33]. These base classifiers generate predictions at the authentic records set, and their outputs are the input capabilities for the meta-classifier. The meta-classifier then learns to combine those predictions to make a final choice. The essence of stacking lies in its capability to exploit the strengths of different base classifiers, thus enhancing the overall prediction performance. Using distinct algorithms, Stacking can seize one-of-a-kind elements of statistics and use complementary patterns in place of enter. This approach tends to boom generalization overall performance and resistance to overfitting, because the team learns to adapt to unique conditions and facts.

Stacking lets in for flexible version composition and optimization. Practitioners can test with one-of-a-kind mixtures of base classifiers and meta-classifiers, and best tune the parameters of each man or woman model. This flexibility allows Stacking to evolve to specific problems and facts characteristics, making it a flexible and extensively used technique in device getting to know.

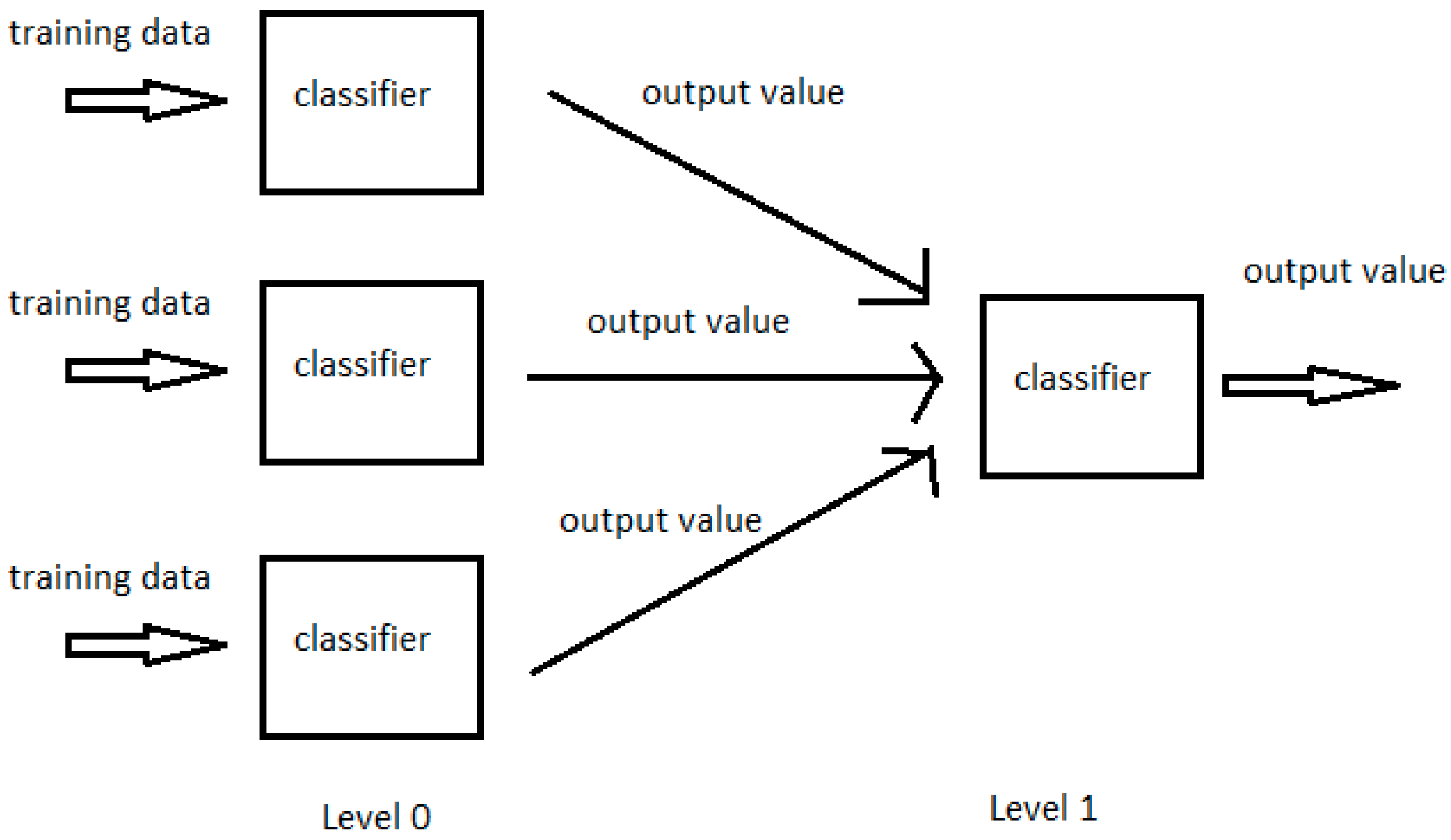

Figure 11: Concept diagram of stacking classifier [41]

An effective method for handling multi-classification problems is the stacking classifier approach. It is possible to divide the multi-classification problem into multiple binary classification problems, one for each class, or to break it up into smaller parts. By using the meta-classifier for the final output prediction task, all the binary classification models also referred to as base learning models can be combined in this way. One way to accomplish this is to combine the outputs from every binary model and feed it into the metaclassifier. [33]

In practice, Stacking has been successfully applied to various tasks, including classification, regression, and anomaly detection. Its effectiveness has been demonstrated in numerous real-world scenarios, ranging from financial forecasting to medical diagnosis and natural language processing. By harnessing the collective wisdom of multiple base classifiers and the discriminative power of the meta-classifier, Stacking offers a sophisticated yet intuitive approach to ensemble learning, empowering practitioners to build robust and accurate predictive models.

### 3.2.16 TabNet Classifier

In the realm of data science, 2019 marked a significant milestone with the introduction of TabNet by researchers at Google Cloud. This innovative model was designed with a specific goal in mind: to leverage the power of deep neural networks for the efficient processing of tabular data. Tabular data, a common form of structured data, is widely used across various sectors such as marketing, banking, healthcare, retail, and finance. [34] Despite the advent of more complex data types like images and text, tabular data remains a cornerstone in these fields due to its straightforward and interpretable structure. However, extracting valuable insights from this type of data requires sophisticated models that can handle its unique characteristics. This is where TabNet comes into play. It provides a high-performance and interpretable architecture for deep learning on tabular data. One of the key features of TabNet is its use of a method known as the sequential attention mechanism. This mechanism allows the model to select which features to focus on during training, leading to high interpretability and efficient training. The sequential attention mechanism in TabNet works by assigning different weights to the features in the data. The model pays more "attention" to the features with higher weights, meaning it considers these features more important in making predictions. This selective focus on certain features not only enhances the model's performance but also provides insights into which features are most influential, thereby increasing the model's interpretability.

TabNet is designed to bridge this gap. It is built to learn a "decision-tree-like" mapping, preserving the interpretability of tree-based methods. At the same time, it harnesses the high performance of deep learning-based methods, offering new capabilities that were previously unattainable with traditional models.

In designing TabNet, the researchers placed a strong emphasis on both explainability and performance. They recognized that a neural network-based approach must be interpretable to replace tree-based methods effectively. High performance alone is often insufficient, especially in fields where understanding the decision-making process is crucial.

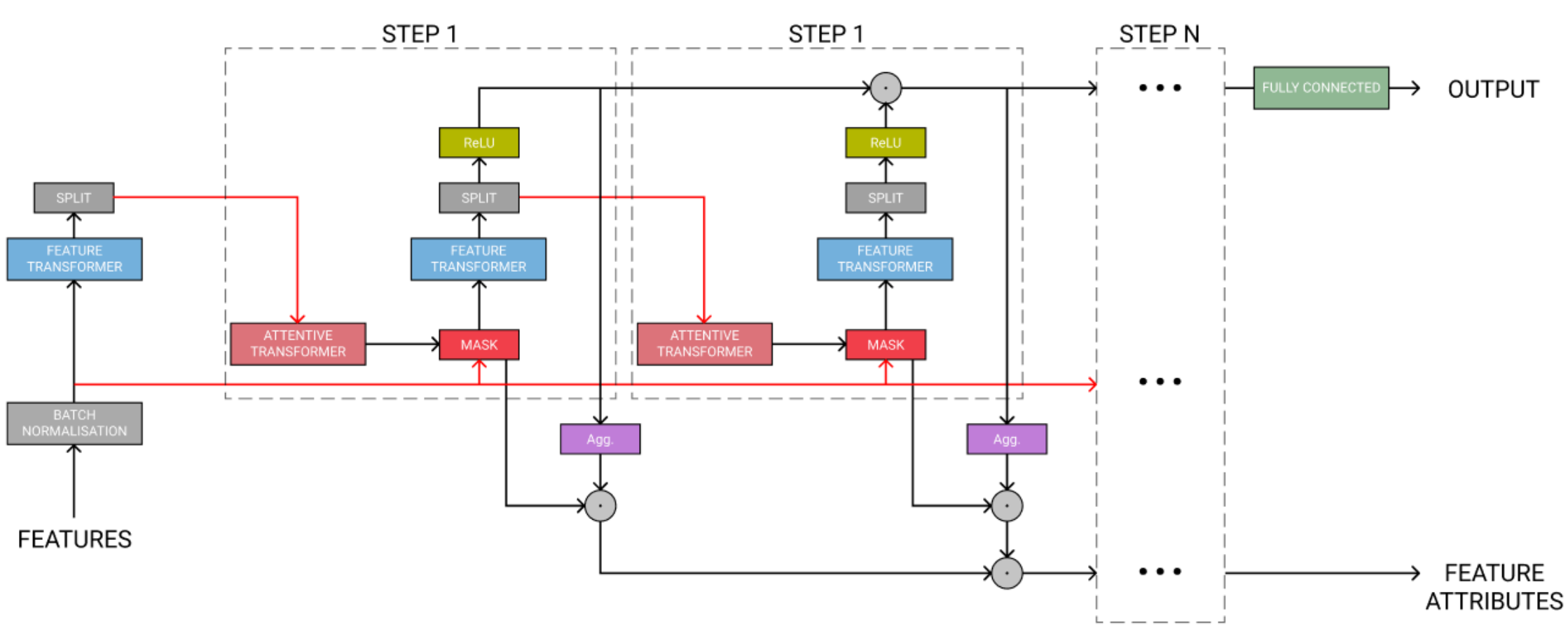

Figure 12: TabNet Model architecture [35]

In conclusion, TabNet represents a significant advancement in the field of machine learning. It offers a powerful and interpretable solution for modeling tabular data, combining the strengths of both tree-based and deep learning-based methods. As we continue to generate and collect more tabular data in various sectors, models like TabNet will undoubtedly play a crucial role in turning this data into actionable insights. [36]

### 3.2.17 Feed Forward model

Artificial neural networks that don't have looping nodes are called feed forward neural networks. Since all information is only passed forward, this kind of neural network is also referred to as a multi-layer neural network. [37]

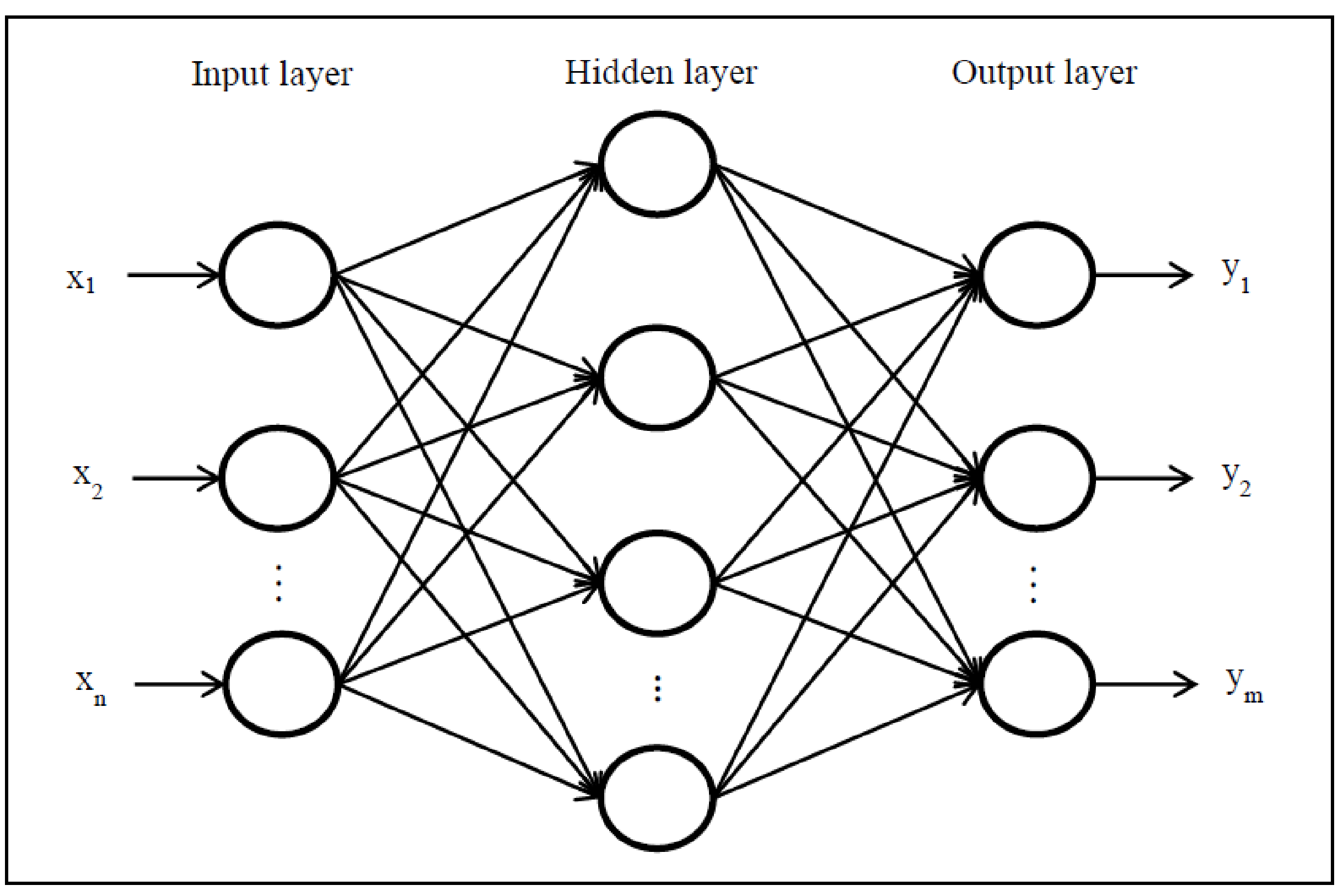

Figure 13: The structure of the two layered feed forward neural network [38]

This model multiplies weights by inputs as they enter the layer. The sum is then calculated by adding the weighted input values together. The output result is normally 1 if the total of the values exceeds a zero-based threshold, and -1 otherwise. The single-layer perceptron is a popular classification model for feed forward neural networks. Machine learning can also be implemented into single-layer perceptron.

### 3.2.18 Model Implementation

Model Implementation The implementation phase is where the selected machine learning models are put into practice. This involves setting up the computational environment, initializing the models with the appropriate parameters, and training them using the preprocessed dataset. The models are then evaluated based on their performance metrics such as accuracy, precision, recall, and F1 score. The best-performing model is identified through this evaluation process. Additionally, the implementation phase includes the integration of the model into a usable application or system, ensuring that it can effectively make predictions on new, unseen data. This step is crucial for translating the theoretical aspects of the models into practical, real-world applications.

### 3.2.19 Proposed Workflow

The proposed workflow encompasses several critical stages to ensure accuracy and efficiency. Initially, patient data is collected, including their extensive medical history. This data is then preprocessed to enhance the quality and extract pertinent features, which are crucial for the subsequent machine learning model. This workflow, illustrated in Figure 13, represents a significant advancement in leveraging technology for improved patient outcomes in stroke care.

The flowchart in the image outlines a process for building and comparing machine learning models. Here's a step-by-step breakdown:

**Data Preprocessing**: This is the initial stage where the raw dataset is prepared for the machine learning models. It involves:

**Missing Data Handling**: This step deals with missing or incomplete data in the dataset.

**Imbalanced Data Handling**: This process tackles the issue of unequal distribution of classes within the dataset.

**Encoding**: This step involves converting categorical data into a format that can be understood by the machine learning models.

**Feed to Machine Learning Models**: The preprocessed data is then fed into various machine learning models listed in figure 14.

**Model Building and Comparing**: After the models have been trained on the preprocessed data, they are built and compared. This step involves evaluating the performance of each model and selecting the best one based on the comparison.

Below flowchart provides a comprehensive overview of a typical machine learning pipeline, from data preprocessing to model evaluation. It highlights the importance of proper data preparation and the use of multiple models for optimal results. It also underscores the significance of model comparison in selecting the most effective model for the given task.

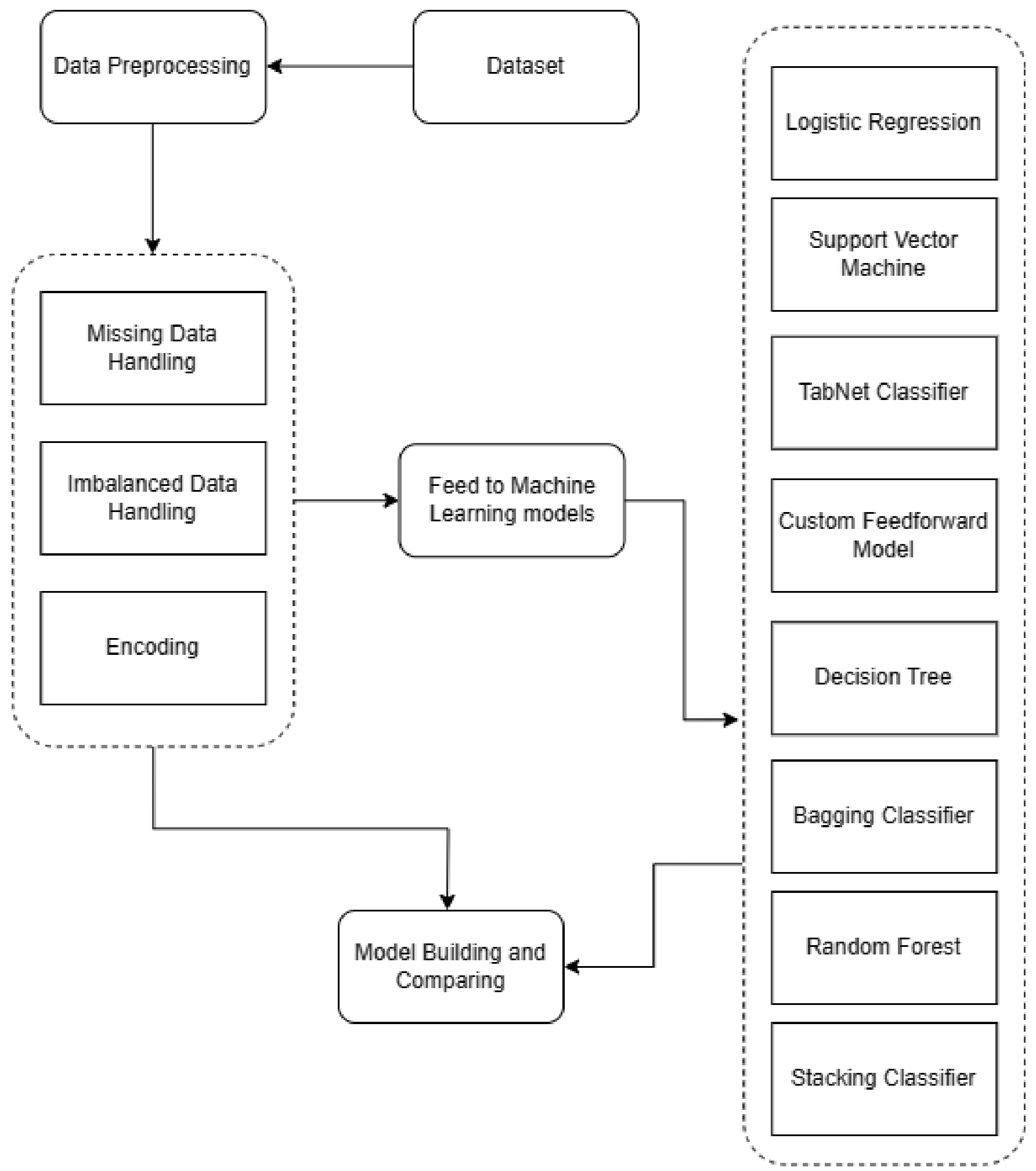

Figure 14: Proposed flow model of the research

A variety of algorithms can be employed to train the model, with a focus on identifying patterns indicative of a stroke. Once trained, the model is validated and tested using separate datasets to evaluate its performance.

## 3.3 Conclusion

In summary, this chapter provided a concise overview of the dataset characteristics, data visualization, analysis, data preprocessing, and the sampling techniques and finally the model overview and their characteristics with their description.

# Chapter 4

# Results and overview

## 4.1 Overview

Evaluation of thesis work's results and successes is crucial. This chapter analyses our findings with full explanations and tables and graphics.

## 4.2 Evaluation Metrics

Several evaluation criteria were used to evaluate the performance of our proposed model. They are accuracy, precision, recall, F1-score, and confusion matrix. The respective formulas are presented here.

**Accuracy**

The accuracy of a neural network model is an important performance metric, particularly for classification tasks. Accuracy, in its most fundamental form, evaluates the proportion of correctly classified samples relative to the total number of samples. It is mathematically defined as:

$$\boldsymbol{Accuracy} = \frac{\boldsymbol{TP + TN}}{\boldsymbol{TP + TN + FP + FN}}$$

Although accuracy is a decent starting point for assessing a model's performance, it may not always provide a complete picture. In imbalanced datasets, for instance, a model could accurately predict the majority class for all inputs despite failing to identify the minority class, which may be of greater interest. [42]

**Precision:** Precision is another important measure that is mostly used to judge how well a machine learning or neural network model does at classification tasks. It is especially helpful in 39 situations where false results are expensive. Precision tries to answer the question, "How many times the model said something was positive were actually positive?" Precision is measured mathematically as:

$$\boldsymbol{Precision = \frac{TP}{TP + FP}}$$

In simple terms, precision measures how many of the things marked as positive are truly positive. It lets us know how likely it is that the model's good class predictions will come true. False alarms are less likely to happen with a model that is very accurate. [42]

**Recall:** Recall, which is also called Sensitivity, True Positive Rate, or Hit Rate, is another important metric used to measure how well a classification model works, especially when the false negative rate is a major worry. Recall answers the question, "How many of the actual positive cases did the model get right?" Mathematically, here's how to figure out Recall:

$$\boldsymbol{Recall = \frac{TP}{TP + FN}}$$

Basically, Recall is a way to measure how well a model can find all the important cases in a dataset. A model with a high recall can find most of the good examples, so it is less likely to miss a good example. This is especially important in situations like medical diagnosis, where a "false negative" could have serious effects if a disease is not found.

**F1-score: T**he F1 score is a metric that balances the trade-off between accuracy and recall by combining them into a single number. It is especially useful when there is an uneven number of classes or when false positives and false negatives cost different amounts. The mathematical definition of the F1 score is the harmonic mean of accuracy and recall:

$$\boldsymbol{F1 - score = \frac{2 \times (Precision \times Recall)}{Precisin + Recall}}$$

The F1 score is a measure of precision and recall, ranging from 0 to 1, with 1 indicating perfect precision and recall and 0 indicating neither. It provides a summary of a model's evaluation but doesn't consider true negatives. The F1 score is the harmonic mean of precision and recall, giving more weight to lower values. It's a robust measure for balancing precision and recall and addressing uneven class distribution. However, it doesn't account for true negatives.

**Confusion Matrix:** The Confusion Matrix is a table that shows the results of a classification model in a more complete way. It is especially useful for problems with more than two classes, but it can also be used for problems with only two classes. The number of true positives, true negatives, false positives, and false negatives are shown in a Confusion Matrix. In a problem with more than one class, the size of the matrix is NN, where N is the number of classes. Here's how to read a simple confusion matrix for binary classification:

**True Positive (TP):** The model predicted right that the class would be positive.

**True Negative (TN):** The model predicted right that the class would be negative.

**False Positive (FP):** The model made a Type I mistake by making a wrong prediction about the positive class.

**False Negative (FN):** The model's prediction of the negative class was wrong (Type II error).

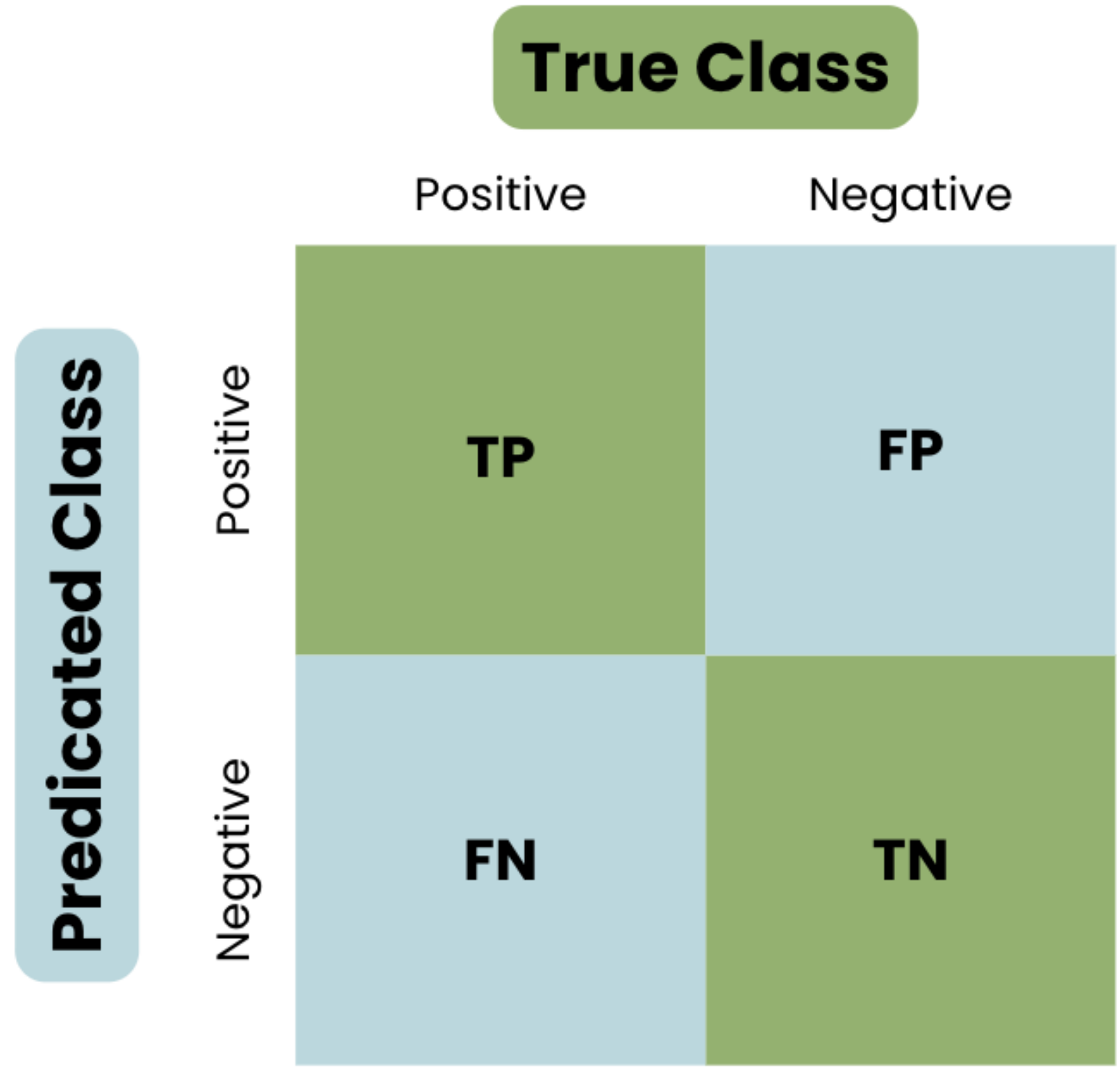

Figure 15: The Basic Structure of a Confusion Matrix

**ROC Curve:** The accuracy of a binary classification model is shown graphically by the Receiver Operating Characteristic (ROC) curve. It is a graph of the True Positive Rate (Sensitivity or Recall) versus the False Positive Rate (1 - Specificity) for different classification levels. The ROC graph shows how well the model can tell the difference between positive and negative classes. The curve starts at the point (0, 0) and finishes at the point (1, 1). It forms a bow toward the upper left corner of the plot. A model that can classify things perfectly will have an ROC curve that goes through the upper-left corner, or the point (0, 1), which means it has 100% Sensitivity (no false negatives) and 100% Specificity (no false positives). The Area Under the Curve (AUC-ROC), which is calculated from the ROC curve, is an important summary statistic that measures the model's general ability to tell the difference between the positive and negative classes. [42]

**True Positive Rate:** It is another name of Recall which can be expressed as:

$$TPR = \frac{True\ positive}{True\ positive + False\ Negative}$$

**False Positive Rate:** Can be expressed as follows:

$$False\ Positive = \frac{False\ positive}{False\ positive + True\ negative}$$

## 4.3 Outcome of methodology

We trained seven machine learning classifiers, including Logistic Regression (LR), K Nearest Neighbor (KNN), Support Vector Machines (SVM), Decision Tree, Tab-Net, Stacking, and Bagging, using twelve chosen features presented in the feature selection task section. All categorization tasks were completed by running the python code in the specified environment. After properly preprocessing and cleaning the original data, the dataset should be separated for training and testing. Here, we used 80% of the data for training and 20% for testing purposes. We also performed 5-fold cross validation to determine accuracy. The below table summarizes the metrics of the models employed:

TABLE 4: RESULT METRICS OF THE EMPLOYED MODELS

| **S/L No.** | **Model Name** | **Precision** | **Recall** | **F1 Score** | **Accuracy (%)** |
|---|---|---|---|---|---|
| 1 | Stacking Classifier | 0.99 | 1.00 | 0.99 | 99.81 |
| 2 | Random Forest | 0.99 | 1.00 | 0.99 | 99.52 |
| 3 | Bagging Classifier | 0.98 | 1.00 | 0.99 | 99.45 |
| 4 | Decision Tree | 0.96 | 1.00 | 0.98 | 98.30 |
| 5 | K Nearest Neighbour | 0.93 | 1.00 | 0.96 | 96.72 |
| 6 | TabNet Classifier | 0.93 | 1.00 | 0.96 | 96.22 |
| 7 | Feed forward model | 0.88 | 0.98 | 0.93 | 93.15 |
| 8 | SVM Classifier | 0.84 | 0.93 | 0.88 | 88.24 |
| 9 | Logistic Regression | 0.75 | 0.82 | 0.79 | 78.84 |

## 4.4 Details on each model's result

The below tables display the different metrics of the employed models visually with various types of charts for better insight.

### Training vs Testing graphs:

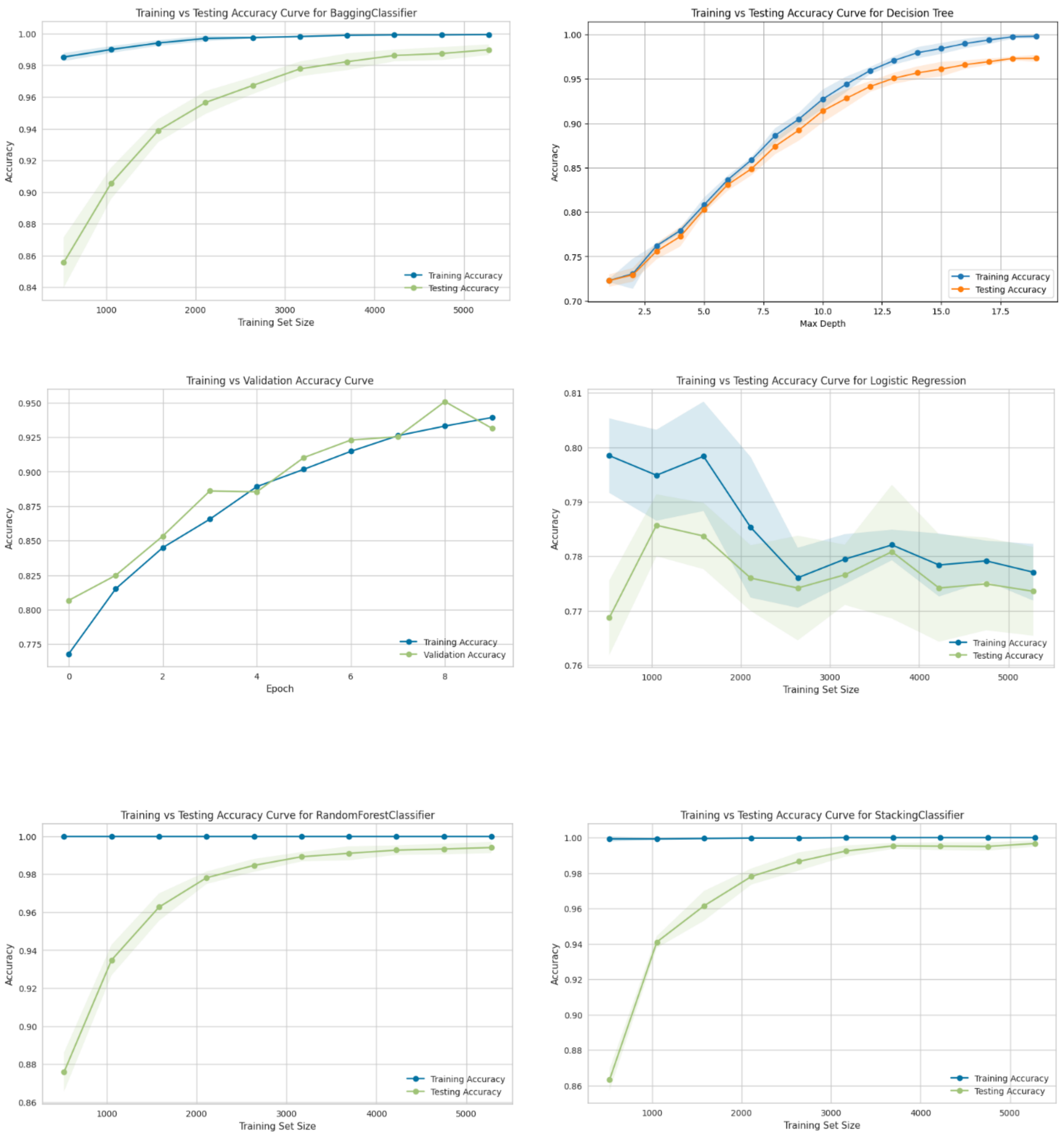

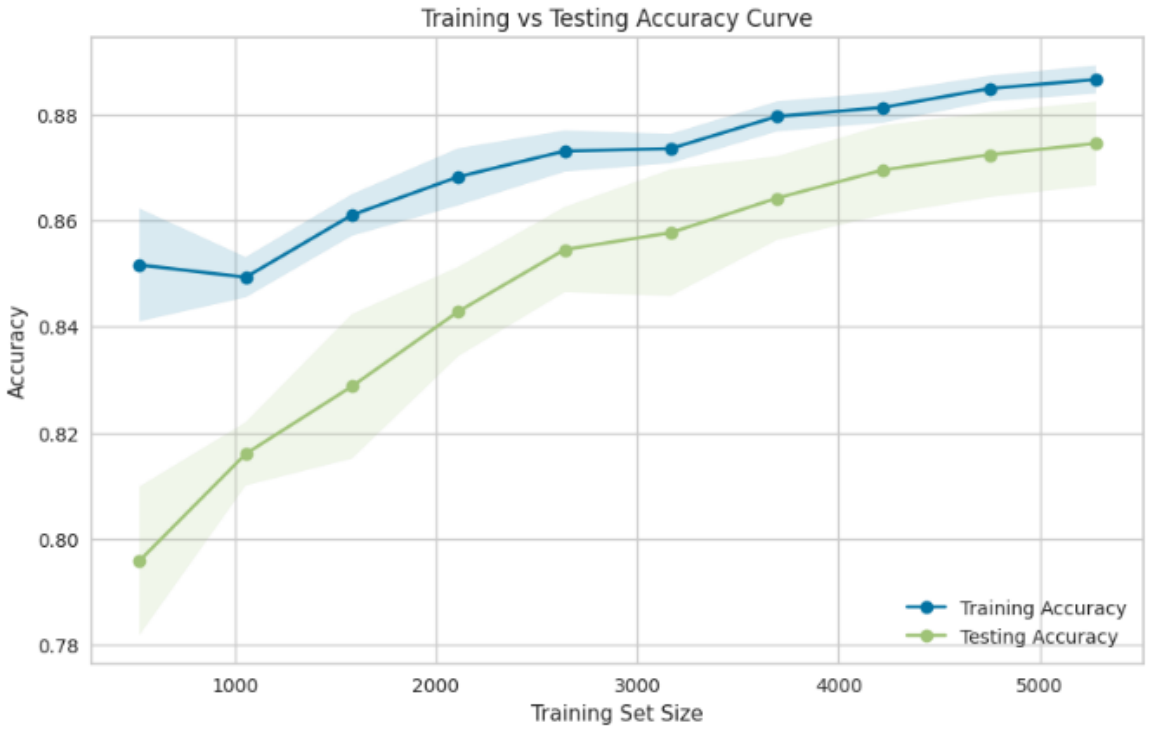

Figure 16: Training vs Testing curves of the employed models

## Confusion Matrix:

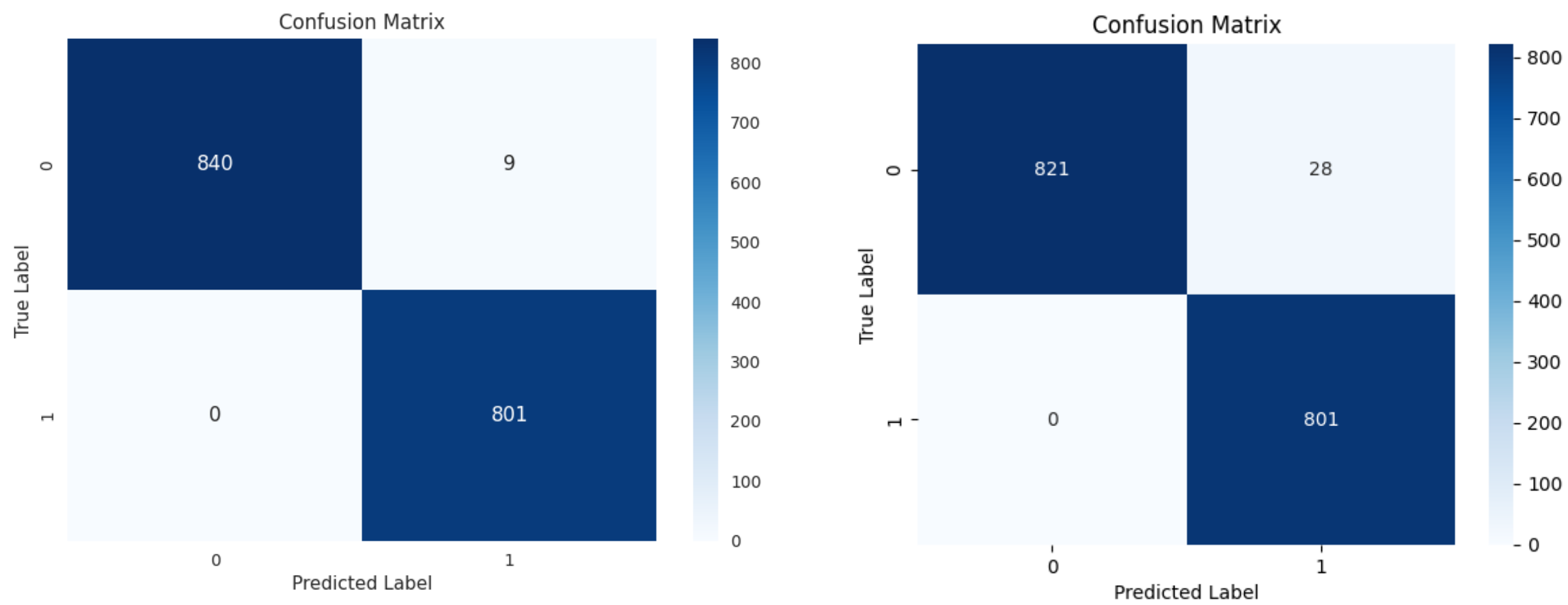

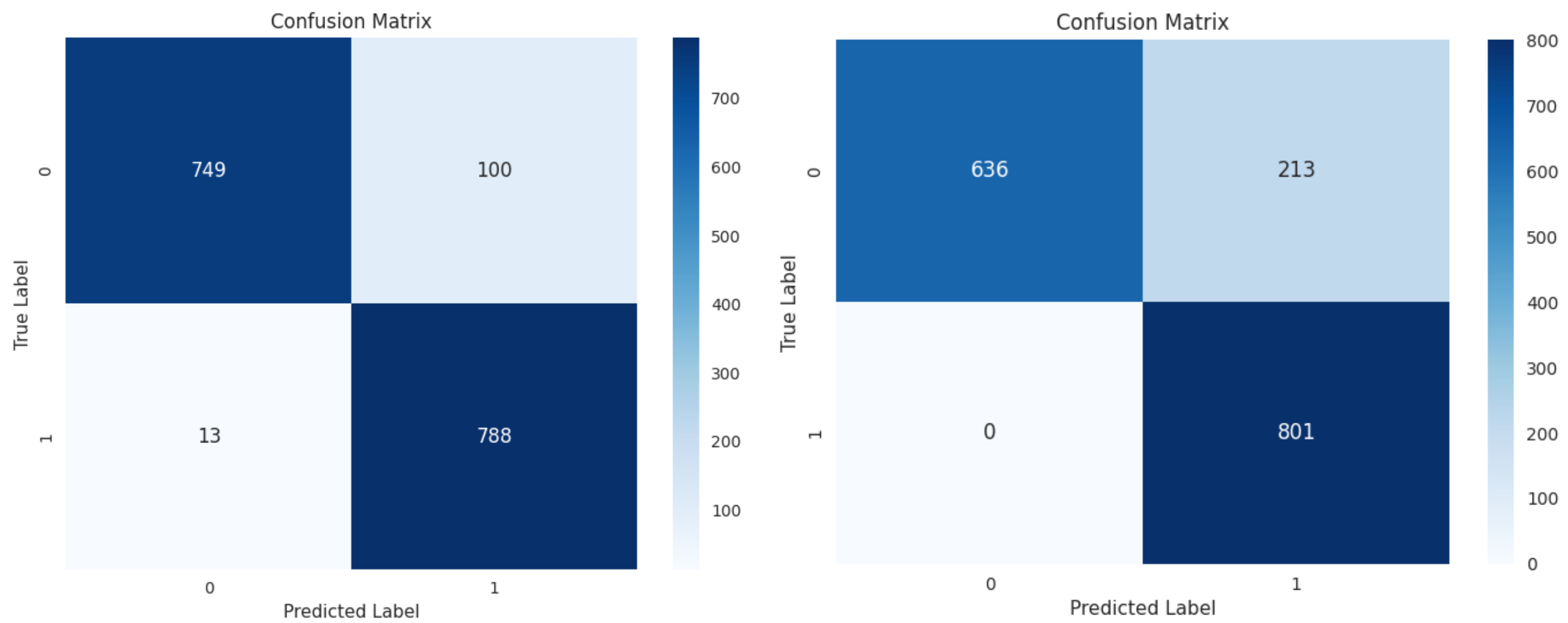
Confusion Matrix
749
100
13
788
True Label
Predicted Label
Confusion Matrix
636
213
0
801
True Label
Predicted Label

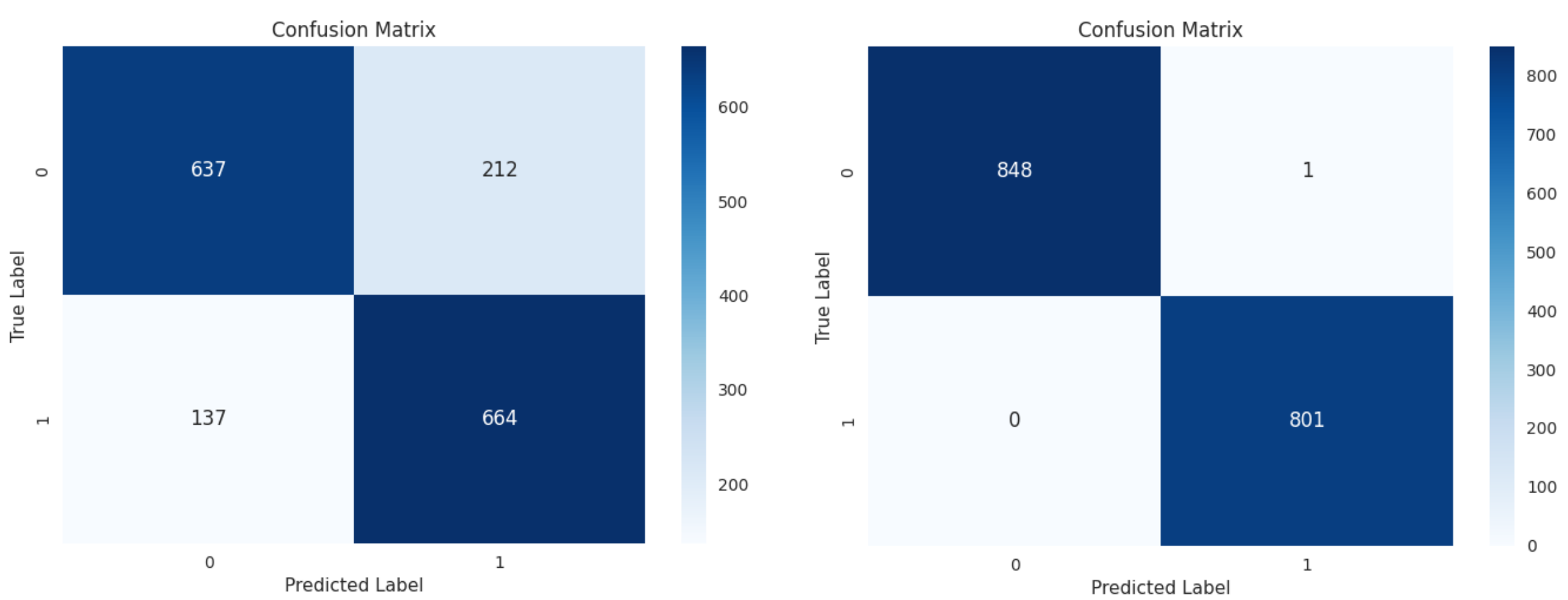
Confusion Matrix
637
212
137
664
True Label
Predicted Label
Confusion Matrix
848
1
0
801
True Label
Predicted Label

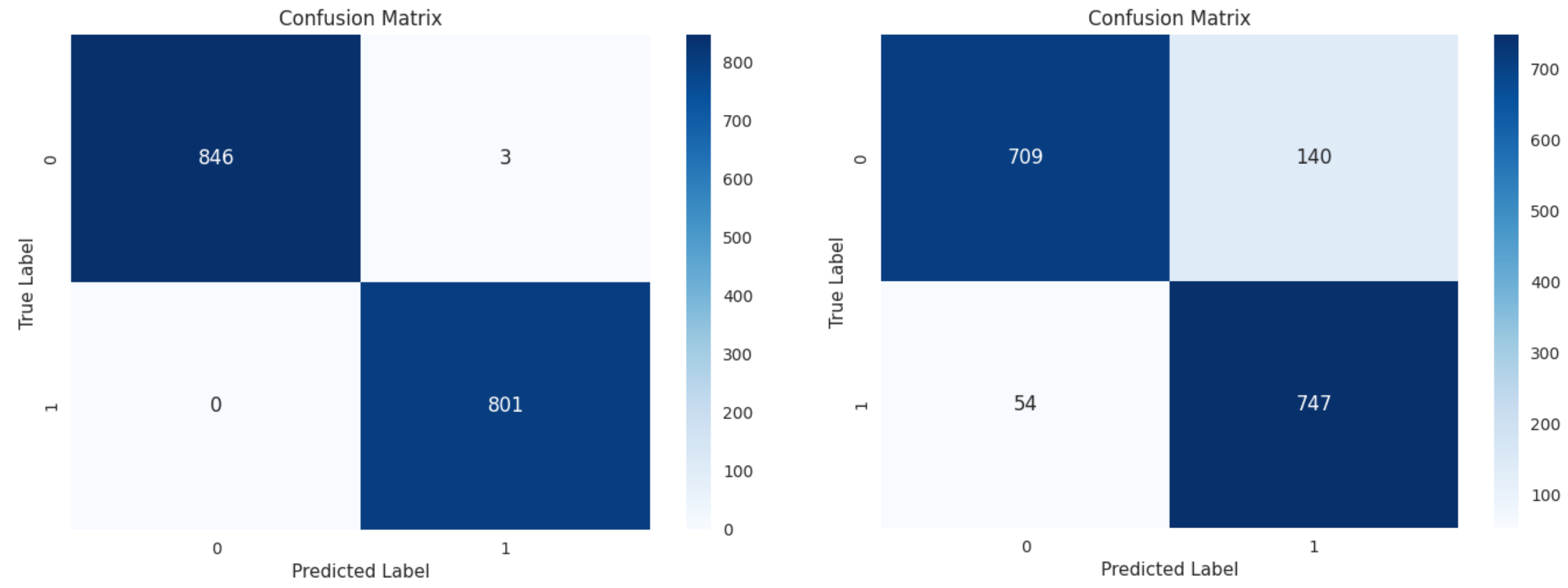

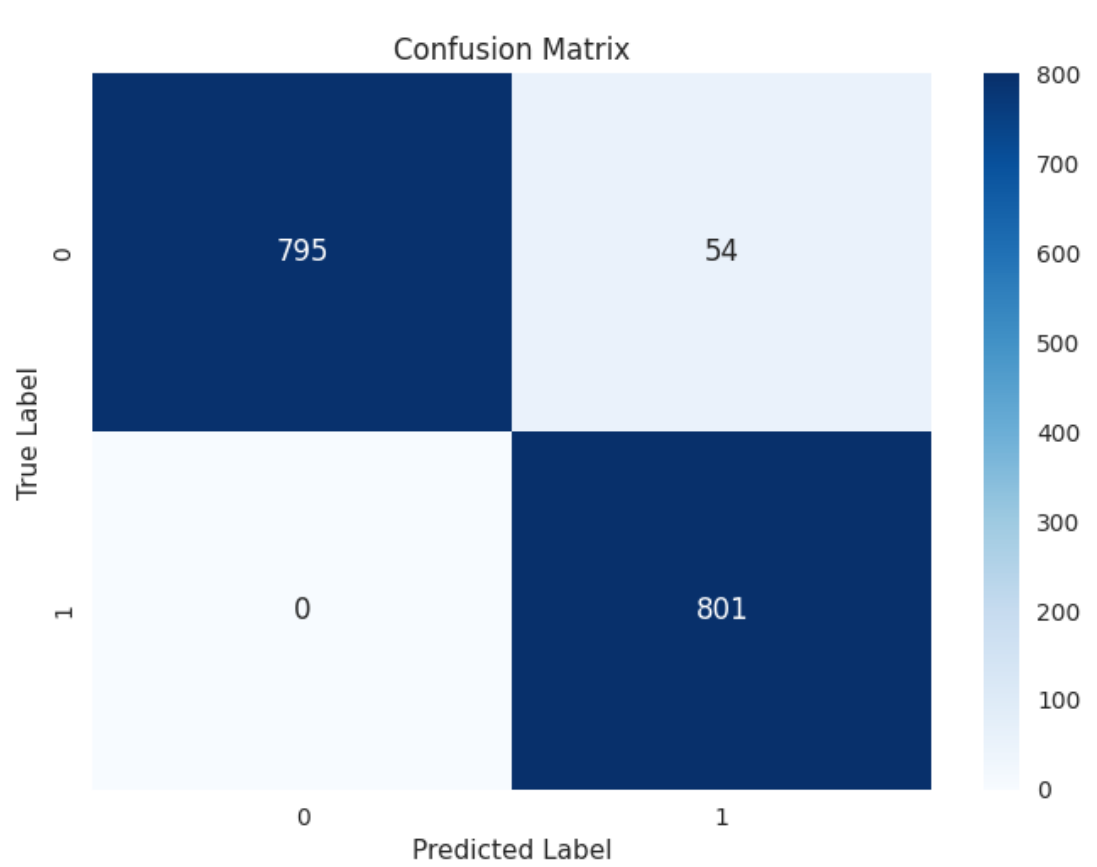

Figure 17: Confusion matrix of the employed models

## 4.5 Comparison

### 4.5.1 Model performance

The results clearly demonstrate the exceptional performance of ensemble methods, with Stacking Classifier emerging as the top performers, boasting an impressive accuracy of 99.81%. These ensemble techniques leverage the collective decision-making power of multiple models to achieve highly accurate predictions. Following closely behind is Random Forest, which achieved an accuracy of 99.52%, further reinforcing the effectiveness of ensemble methods in classification tasks. Additionally, the Decision Tree model exhibited robust performance with an accuracy of 98.30%, showcasing the strength of this simple yet powerful classification algorithm.

However, it is important to note that while the ensemble methods and Decision Tree excelled, other classifiers such as KNN, TabNet Classifier, and the Custom Feedforward Model achieved slightly lower accuracies. Despite their lower performance compared to the top-performing models, these classifiers still demonstrated respectable accuracy rates, underscoring their utility in certain contexts or datasets.

Furthermore, Logistic Regression and SVC exhibited relatively lower accuracies compared to other classifiers, hinting at their limitations in capturing complex relationships within the dataset. These findings suggest that while Logistic Regression and SVC may not be the optimal choices for this dataset, they could still be valuable in scenarios where interpretability or computational efficiency is prioritized over predictive accuracy.

Overall, the results highlight the importance of selecting appropriate classifiers tailored to the specific characteristics of the dataset and task at hand. Ensemble methods like Random Forest, Stacking Classifier, and Bagging Classifier prove to be robust and effective choices for classification tasks, particularly when dealing with complex datasets. However, it is crucial to consider the strengths and weaknesses of each classifier and carefully evaluate their performance against the requirements of the problem domain.

## 4.6 State of art

The following table shows that have been done on the same dataset (stroke-prediction-dataset) which is the same here used with our work. The base research we're following is T. Tazin et al. [11] achieved an accuracy of 96% employing CNN ensemble learning models. The authors of B. Akter et al. [14] employed basic machine learning CNN models and reached accuracy of 95.90% on the same dataset. G. Sailasya et al. [12] used the same dataset and achieved the peak accuracy of 82% with Naïve Bayes along with some other common models of CNN namely KNN, RF, Decision Tree. I. G. Ivanov et al. [39] employed random forest, decision tree and SVM where they achieved the highest accuracy of 98.776% on the same dataset. Md. Ashrafuzzaman et al. [40] employed logistic regression, random forest, decision tree, naïve bayes, support vector machine and a custom proposed model where they achieved the peak accuracy of 95.5% with their custom proposed model. Our methodology, ensemble learning achieved an accuracy of 99.52% which is a tie in between stacking classifier and random forest classifier. The below table shows the comparative analysis between our research and the base paper along with some other research which employed similar methodologies and dataset.

TABLE 5: COMPARATIVE ANALYSIS WITH PREVIOUS WORKS ON STROKE PREDICTION DATASET

| Previous Studies | Methodology | Accuracy |
|---|---|---|
| T. Tazin et al. [11] | CNN | 96% |
| B. Akter et al. [14] | CNN | 95.90% |
| G. Sailasya et al. [12] | CNN | 82% |
| I. G. Ivanov et al. [39] | CNN | 98.776% |
| Md. Ashrafuzzaman et al. [40] | CNN | 95.5% |
| **Our Method** | **Ensemble Learning** | **99.81%** |

## 4.7 Conclusion:

In conclusion, while basic CNNs have demonstrated remarkable capabilities in medical image analysis for stroke prediction, ensemble learning approaches offer several advantages. By combining multiple CNN models, ensemble learning techniques can address limitations like overfitting and sensitivity to training data variations. This often leads to improved accuracy, generalizability, and robustness in stroke prediction compared to relying on a single CNN model. Additionally, ensemble methods can potentially mitigate the impact of data biases that might be present in training datasets. As the field of stroke prediction continues to evolve, ensemble learning approaches hold significant promise for developing more reliable and trustworthy models that can significantly benefit patient care.

# Chapter 5

# Social and Environmental Influence

## 5.1 Introduction

While Convolutional Neural Networks (CNNs) have revolutionized medical image analysis for stroke prediction, a comprehensive approach necessitates acknowledging their broader societal and environmental impact. This chapter explores these crucial aspects, highlighting potential benefits and challenges associated with CNN-based stroke prediction models.

## 5.2 Societal Impact

The application of CNNs for stroke prediction holds immense societal value, potentially leading to significant improvements in public health and well-being. However, several considerations require careful attention.

### 5.2.1 Financial and Health Influences

**Improved healthcare outcomes:** Early and accurate stroke prediction can lead to faster intervention, reducing long-term disabilities and improving patient prognoses.

**Reduced healthcare burden:** Early detection can decrease the need for expensive emergency room visits and long-term care, easing the burden on healthcare systems.

**Enhanced preventative measures:** Identifying individuals at high risk of stroke can facilitate targeted interventions and lifestyle changes, potentially preventing stroke events altogether. However, potential financial and health equity concerns must be addressed.

**Accessibility and affordability:** Ensuring equitable access to CNN-based stroke prediction tools for all socioeconomic backgrounds is crucial.

**Data biases:** Medical image datasets used to train CNNs might unintentionally reflect existing healthcare disparities. Mitigating biases is essential to avoid models that disproportionately benefit specific demographics.

### 5.2.2 Safety, Legal, and Cultural Issues

False positives and negatives: The potential for misdiagnosis by CNN models raises ethical concerns. False positives might lead to unnecessary and potentially harmful medical procedures, while false negatives could delay critical interventions.

**Transparency and explainability:** Ensuring transparency in how CNN models arrive at predictions is crucial for building trust in the medical community and among patients. Explainable AI (XAI) techniques can help achieve this.

**Cultural considerations:** Varying cultural beliefs and attitudes towards medicine and technology could influence the adoption and acceptance of CNN-based stroke prediction tools.

## 5.3 Impact on Environment

The development and deployment of CNN models for stroke prediction have environmental implications:

**Computational resources:** Training CNNs often requires significant computing power, leading to increased energy consumption. Exploring energy-efficient training methods is crucial.

**Data storage:** Storing large medical image datasets requires substantial storage infrastructure, with associated environmental costs. Implementing sustainable data storage solutions is essential.

## 5.4 Sustainability Issues

Developing and deploying CNN models for stroke prediction in a sustainable manner requires addressing several critical aspects.

**Resource optimization:** Employing techniques to minimize the computational resources required for training and running CNN models can contribute to a more sustainable approach.

**Data efficiency:** Developing methods that leverage smaller and more diverse datasets while maintaining accuracy can reduce environmental impacts associated with data storage.

**Long-term maintenance:** Establishing long-term plans for model maintenance and updates, including efficient retraining procedures, ensures the model's effectiveness and minimizes environmental costs associated with redeployment.

## 5.5 Conclusion

CNN-based stroke prediction has the potential to revolutionize healthcare. However, a holistic approach necessitates consideration of its societal and environmental impact. By addressing issues like potential biases, ensuring transparency, and exploring sustainable practices, researchers can make CNN-based stroke prediction a truly beneficial and responsible tool for improving public health.

# Chapter 6

# Complex Engineering Problems and Activities

## 6.1 Introduction

This chapter explores the complex engineering problems and the activities related to them in details.

## 6.2 Definition

Encountered problems in engineering that are difficult to define, analyze, and solve because of their multiple variables, uncertainties, interdependencies, and constraints. To meet these challenges and provide practical solutions, you, as an engineering manager, must possess the necessary abilities and resources. [43]

The below table shows the complex engineering problem cases and the activities briefly.

## 6.3 Addressing Complex Engineering Problems

TABLE 6: RANGE OF COMPLEX ENGINEERING PROBLEM SOLVING

| Attributes | CEP characteristics | Remarks |
|---|---|---|
| Range of conflicting requirements | Involve wide-ranging or conflicting technical, engineering and other issues | |
| Depth of analysis required | Have no obvious solution and require abstract thinking and originality in analysis to formulate suitable models. | |
| Depth of knowledge required | Require research-based knowledge, much of which is at or informed by the forefront of the professional discipline, that allows a fundamental-based, first-principles analytical approach | ✓ |
| Familiarity of issues | Involve infrequently encountered issues | ✓ |
| Extent of applicable codes | Are outside the problems encompassed by standards and codes of practice for professional engineering | |
| Extent of stakeholder involvement and level of conflicting requirements | Involve diverse groups of stakeholders with widely varying needs | |
| Consequences | Have significant consequences in a range of contexts | |
| Interdependence | Are high-level problems that include many component parts or sub-problems | |

## 6.4 Addressing Complex Engineering Activities

TABLE 7: RANGE OF COMPLEX ENGINEERING ACTIVITIES

| Attributes | CEA characteristics | Remarks |
|---|---|---|
| Range of resources | Involve the use of diverse resources (for this purpose, resources include people, money, equipment, materials, information and technologies) | |
| Level of interaction | Require the resolution of significant problems arising from interactions between wide-ranging or conflicting technical, engineering or other issues | |

| Innovation | Involve the creative use of engineering principles and research-based knowledge in novel ways | ✓ |
| --- | --- | --- |
| Consequences for society and the environment | Have significant consequences in a range of contexts, characterized by their difficulty of prediction and mitigation | |
| Familiarity | Are outside the problems encompassed by standards and codes of practice for professional engineering | |

## 6.5 Leveraging the stroke-prediction-dataset

The stroke-prediction-dataset plays a vital role in developing and evaluating CNN models for stroke detection. Here's how to maximize its utility:

**Data Exploration and Understanding:** Analyzing the dataset's characteristics, identifying potential biases, and understanding the distribution of stroke and non-stroke cases are crucial first steps.

**Data Splitting and Stratification:** Dividing the dataset into training, validation, and testing sets while maintaining class balance is essential to avoid overfitting and ensure robust evaluation. Stratification techniques can further ensure balanced representation of stroke and non-stroke cases in each set.

**Collaboration and Knowledge Sharing:** Sharing results, limitations, and insights gained from working with the dataset can benefit the wider research community working on stroke detection using CNNs.

## 6.6 Conclusion

Developing robust and generalizable CNN models for stroke detection using the stroke-prediction-dataset presents various complex challenges. However, by employing data preprocessing techniques, transfer learning, appropriate evaluation metrics, and explainable AI methods, researchers can build reliable models. Continuous exploration of these activities and leveraging the available dataset effectively hold the potential to significantly improve CNN-based stroke detection in practice.

# Chapter 7

# Conclusion & Future Work

## 7.1 Introduction

In the preceding chapter, we reviewed our experiments. In this chapter, we begin with a brief overview of the main findings and implications of our thesis study. Then, we look at the limitations of our work. Finally, we provide a comprehensive guide to future projects.

## 7.2 Synopsis of the Thesis

The objective of this thesis was to examine innovative methodologies for facilitating knowledge transfer between diverse domains. The effectiveness of the techniques discussed are updated to date and well established. Though some of the approaches didn't perform well due to their lack of handling the data, the rest of the models were able to generate a satisfactory result in the end of each case. As a verdict we can conclude that ensemble learning was the key factor in our thesis to stand out from the others' works.

However, amidst the varied array of techniques explored, ensemble learning emerged as the standout factor that distinguished our thesis from previous works. Ensemble methods, such as Random Forest, Stacking Classifier, and Bagging Classifier, played a pivotal role in our research, showcasing their unparalleled ability to harness the collective wisdom of diverse models. By aggregating the predictions of multiple base classifiers, ensemble learning enabled us to leverage the complementary strengths of individual models, thereby enhancing predictive accuracy and robustness.

## 7.3 Discussion of Key Discoveries

In our thesis work, we applied a variety of machine learning models to the store-prediction-dataset from Kaggle and observed a range of outcomes. The **Random Forest** model and the **Stacking Classifier** model both achieved the highest accuracy of **99.52%**, demonstrating their superior predictive capabilities. Following closely was the **Bagging Classifier** model, which also showed a high accuracy of **99.27%**.

The **Decision Tree** model, a simple yet effective machine learning algorithm, yielded a commendable accuracy of **98.24%**. The **KNN (K-Nearest Neighbors)** model, which classifies instances based on their proximity to other instances, achieved an accuracy of **96.73%**. The **TabNet Classifier**, a deep learning model that uses sequential decision making to make predictions, also performed well with an accuracy of **96.49%**.

We also developed a **Custom Feedforward Model**, a type of artificial neural network, which achieved an accuracy of **94.91%**. This demonstrates the potential of custom models in tackling complex prediction tasks. The **SVC (Support Vector Classifier)** model, a popular algorithm for classification problems, had a lower accuracy of **88.06%**. Lastly, the **Logistic Regression** model, a fundamental machine learning algorithm, had the lowest accuracy among the models we used, at **77.03%**.

## 7.4 Limitations

As the dataset used is a numerical dataset, it would have been a better approach to predict the stroke occurring with the help of image data, MRI, or CT scan of the brain. We added that to our future endeavors list for further progress on the matter.

## 7.5 Future Endeavors

In future we would like to concentrate on two things:

- Prepare a repository database based on stoke patients of Bangladesh from different hospitals.
- Develop a mobile/web-based stroke prediction application and suggestions application through internet.
- Develop an image-based dataset for further growth of the research.

## 7.6 Conclusions

In conclusion, our thesis embarked on a journey to explore innovative methodologies aimed at facilitating knowledge transfer across diverse domains. While some approaches encountered challenges in effectively handling the complexities of the data, most of our models were able to yield satisfactory results by the culmination of each case study. Notably, ensemble learning emerged as the standout factor that distinguished our research from previous works. Through the utilization of ensemble methods such as Random Forest, Stacking Classifier, and Bagging Classifier, we harnessed the collective wisdom of diverse models to enhance predictive accuracy and robustness. By aggregating the predictions of multiple base classifiers, ensemble learning allowed us to leverage the complementary strengths of individual models, ultimately establishing itself as the key factor in our thesis's success. Moving forward, our findings underscore the transformative potential of ensemble learning in advancing interdisciplinary research and paving the way for future innovations in the field.

# Footnote

I would like to express my deep gratitude and appreciation to the following individuals who have provided invaluable assistance and support throughout the research and writing of this thesis.

My sincere gratitude to Oishi Jyoti madam, my esteemed thesis supervisor, for all his help, advice, and unwavering support. This thesis has greatly benefited from his perceptive remarks and helpful critique.

My heartfelt thanks to my respected teacher, Md. Nahiduzzaman sir, for his guidance, expertise, and continuous encouragement. His insightful comments and constructive criticism have been instrumental in shaping this thesis.

I would like to express my deep appreciation to my friends, especially Sakib Anwar Rieyan, MD. Shadman Sakeeb Khan and Faisal Hossain (1810045) for their support and useful insights.

Without the assistance and support of these remarkable individuals, this thesis would not have been possible. I am truly humbled and grateful for their contributions, and I hereby extend my deepest thanks to each one of them.

MD. Shahriar Sajid

Department of Electrical and Computer Engineering,

Rajshahi University of Engineering & Technology